\documentclass[acmsmall,screen,nonacm=true]{acmart}

\AtBeginDocument{%
  \providecommand\BibTeX{{%
    \normalfont B\kern-0.5em{\scshape i\kern-0.25em b}\kern-0.8em\TeX}}}

\setcopyright{acmcopyright} 
\copyrightyear{2021}
\acmYear{2021}
\acmDOI{10.1145/1122445.1122456}

\acmJournal{JACM}
\acmVolume{xx}
\acmNumber{x}
\acmArticle{}
\acmMonth{7}

\newcommand{\ra}[1]{\renewcommand{\arraystretch}{#1}}
\newcommand\tabitem{\makebox[1.4em][r]{\textbullet~}}
\usepackage{pbox}
\usepackage{enumitem}
\usepackage{graphicx}
\usepackage[]{draftfigure}
\usepackage{tablefootnote}
\usepackage{multirow}
\usepackage{makecell}
\usepackage{comment}

\usepackage{gb4e}
\noautomath 

\usepackage{trfsigns}
\usepackage{amsmath}
\hypersetup{allcolors=blue,colorlinks=true}

\begin{document}

\title{Named Entity Recognition and Classification on Historical Documents: A Survey}


\author{Maud Ehrmann}
\email{maud.ehrmann@epfl.ch}
\orcid{0000-0001-9900-2193}
\affiliation{%
  \institution{Ecole Polytechnique Fédérale de Lausanne}
}

\author{Ahmed Hamdi}
\email{ahmed.hamdi@univ-lr.fr}
\orcid{0000-0002-9283-6969} 
\affiliation{%
  \institution{University of La Rochelle}
}

\author{Elvys Linhares Pontes}
\email{elvys.linhares_pontes@univ-lr.fr}
\orcid{0000-0002-9571-5193}
\affiliation{%
  \institution{University of La Rochelle}
}

\author{Matteo Romanello}
\email{matteo.romanello@epfl.ch}
\orcid{0000-0002-7406-6286}
\affiliation{%
  \institution{Ecole Polytechnique Fédérale de Lausanne}
}

\author{Antoine Doucet}
\email{antoine.doucet@univ-lr.fr}
\orcid{0000-0001-6160-3356}
\affiliation{%
  \institution{University of La Rochelle}
}

\renewcommand{\shortauthors}{Ehrmann et al.}

\begin{abstract}

After decades of massive digitisation, an unprecedented amount of historical documents is available in digital format, along with their machine-readable texts. While this represents a major step forward with respect to preservation and accessibility, it also opens up new opportunities in terms of content mining and the next fundamental challenge is to develop appropriate technologies to efficiently search, retrieve and explore information from this ‘big data of the past’. Among semantic indexing opportunities, the recognition and classification of named entities are in great demand among humanities scholars. Yet, named entity recognition (NER) systems are heavily challenged with diverse, historical and noisy inputs. In this survey, we present the array of challenges posed by historical documents to NER, inventory existing resources, describe the main approaches deployed so far, and identify key priorities for future developments.

\end{abstract}



\maketitle


\section{Introduction}
\label{intro}

For several decades now, digitisation efforts by cultural heritage institutions are 
contributing an increasing amount of facsimiles of historical documents. Initiated in the 1980s with small scale, in-house projects, the `rise of digitisation'  grew further until it reached, already in the early 2000s, a certain maturity with large-scale, industrial-level digitisation campaigns~\cite{terras_rise_2011}. Billions of images are being acquired and, when it comes to textual documents, their content is transcribed either manually via dedicated interfaces, or automatically via optical character recognition (OCR) or handwritten text recognition (HTR)~\cite{causer_many_2014,muehlberger_transforming_2019}. 
As a result, it is nowadays commonplace for memory institutions (e.g. libraries, archives, museums) to provide digital repositories that offer rapid, time- and location-independent access to facsimiles of historical documents as well as, increasingly, full-text search over some of these collections.

Beyond this great achievement in terms of preservation and accessibility, the availability of historical records in machine-readable formats bears the potential of new ways to engage with their contents. In this regard, the application of machine reading to historical documents is potentially transformative, and the next fundamental challenge is to adapt and develop appropriate technologies to efficiently search, retrieve and explore information from this ‘big data of the past’~\cite{kaplan_big_2017}. 
Here research is stepping up and the interdisciplinary efforts of the digital humanities (DH), natural language processing (NLP) and computer vision communities are progressively pushing forward the processing of facsimiles, as well as the extraction, linking and representation of the complex information enclosed in transcriptions of digitised collections. 
In this endeavor, information extraction techniques, and particularly named entity (NE) processing, can 
be considered among the first and most crucial processing steps.

Named entity recognition and classification (NER for short) corresponds to the identification of entities of interest in texts, generally of the types \textit{Person}, \textit{Organisation} and \textit{Location}. Such entities act as referential anchors which underlie the semantics of texts and guide their interpretation. Acknowledged some twenty years ago, NE processing has undergone major evolution since then, from entity recognition and classification to entity disambiguation and linking, and is representative of the evolution of information extraction from a document- to a semantic-centric view point~\cite{rao_entity_2013}. As for most NLP research areas, recent developments around NE processing are dominated by 
deep neural networks 
and the usage of embedded language representations~\cite{collobert_natural_2011,lample_neural_2016}. Since their inception up to now, NE-related tasks are of ever-increasing importance and at the core of virtually any text mining application.

From the NLP perspective, NE processing is useful first and foremost in information retrieval, or the activity of retrieving a specific set of documents within a collection given an input query. Guo et al.~\cite{guo_named_2009} as well as Lin et al.~\cite{lin_active_2012} showed that more than 70\% of queries against modern search engines contain a named entity, and it has been suggested that more than 30\% of content-bearing words in news text correspond to proper names~\cite{gey_research_2000}. Entity-based document indexing is therefore desirable. NEs are also highly beneficial in information extraction, or the activity of finding information within large volumes of unstructured texts. The extraction of salient facts about predefined types of entities in free texts is indeed an essential part of question answering~\cite{molla_named_2006}, media monitoring~\cite{steinberger_introduction_2009}, and opinion mining~\cite{balahur_sentiment_2010}. Besides, NER is helpful in machine translation~\cite{hermjakob_name_2008}, text summarisation~\cite{kabadjov_multilingual_2013}, and document clustering~\cite{escoter_grouping_2017}, especially in a multilingual setting~\cite{steinberger_multilingual_2013a}. 

\begin{figure*}[t]
\centering
    \includegraphics[width=\linewidth]{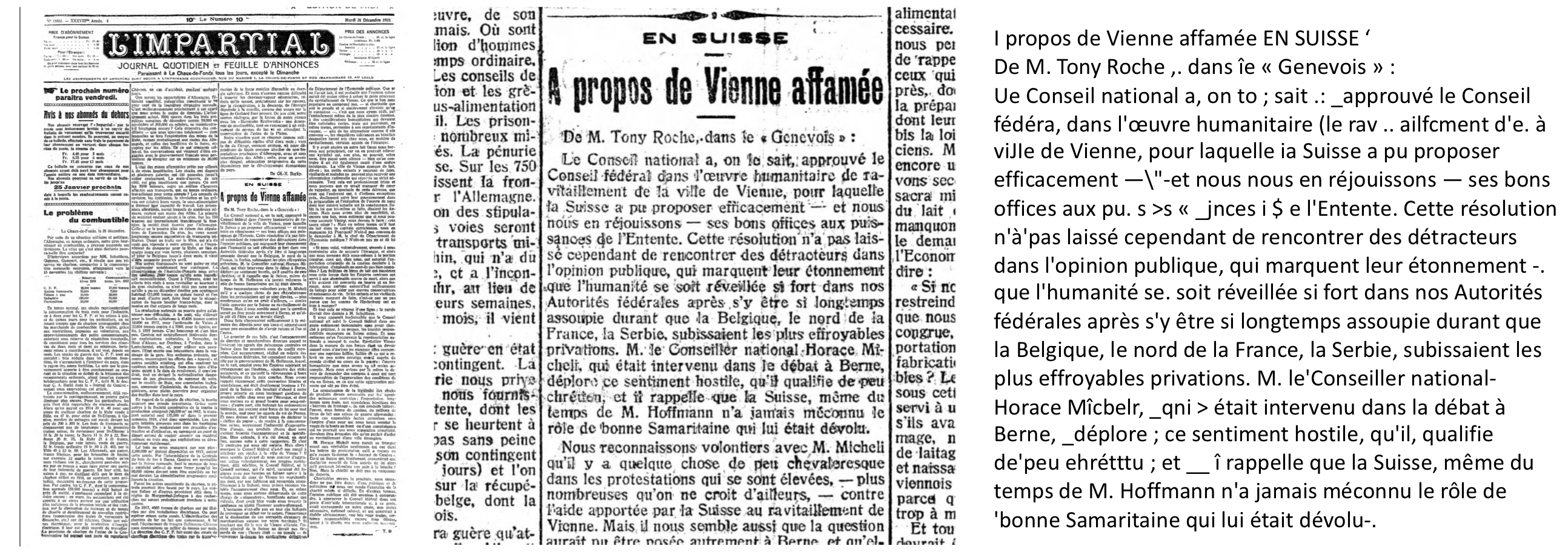} 
    \caption{Swiss journal \textit{L'Impartial}, issue of 31 Dec 1918. Facsimile of the first page (left), zoom on an article (middle), and OCR of this article as provided by the Swiss National Library (completed in the 2010s) (right).}
    \label{fig:newspaper}
\end{figure*}

As for historical material (cf. Figure~\ref{fig:newspaper}), primary needs also revolve around retrieving documents and information, and NE processing is of similar importance~\cite{colavizza_indexdriven_2019}. There are less query logs over historical collections than for the contemporary web, but several studies demonstrate how prevalent entity names are in humanities users’ searches: 80\% of search queries on the national library of France's portal \textit{Gallica} contain a proper name~\cite{chiron_impact_2017}, and geographical and person names dominate the searches of various digital libraries, be they of artworks, domain-specific historical documents, historical newspapers, or broadcasts~\cite{bates_getty_1996,chardonnens_mining_2017,huurnink_search_2010}. 
Along the same line, several user studies emphasise the role of entities in various phases of the information-seeking workflow of historians~\cite{duff_accidentally_2002,gooding_exploring_2016}, 
now also reflected in the `must-have' of exploration interfaces, e.g. as search facets over historical newspapers~\cite{ehrmann_historical_2019,pfanzelter_digital_2021} or as automatic suggestions over large-scale cultural heritage records~\cite{gordea_named_2020}. Besides document indexing, named entity recognition can also benefit downstream processes (e.g. biography reconstruction~\cite{fokkens_proceedings_2018} or event detection~\cite{smith_detecting_2002}), as well as various data analysis and visualisation (e.g. on networks~\cite{vandecamp_link_2011}). Finally, and perhaps most importantly, NER is the first step of entity linking, which can support the cross-linking of multilingual and heterogeneous collections based on authority files and knowledge bases. Overall, entity-based semantic indexing can greatly support the search and exploration of historical documents, and NER is increasingly being applied on such a material.

Yet, the recognition and classification of NEs in historical texts is not straightforward, and performances are rarely on par with what is usually observed on contemporary, well-edited English news material~\cite{ehrmann_diachronic_2016}. In particular, NER on historical documents faces the challenges of domain heterogeneity, input noisiness, dynamics of language, and lack of resources. If some of these issues have already been tackled in isolation in other contexts (with e.g. user-generated text), what makes the task particularly difficult is their combination, as well as their magnitude: texts are severely noisy, domains and time periods are far apart, and there is no (or not yet) historical web to easily crawl to capture language models. In this context of new material, interests and needs, and 
in times of rapid technological change with deep learning, this paper presents a survey of NER research on historical documents. The objectives are to study the main challenges facing named entity recognition and classification when applied to historical documents, to inventory the strategies deployed to deal with them so far, and to identify key priorities for future developments. 

Section~\ref{sec:methodology} outlines the objectives, the scope and the methodology of the survey, and Section~\ref{sec:background} provides background on NE processing. Next, Section~\ref{sec:challenges} introduces and discusses the challenges of NER on historical documents. In response, Section~\ref{sec:resources} proposes an inventory of existing resources, while Section ~\ref{sec:approaches} and ~\ref{sec:strategies} present the main approaches, in general and in view of specific challenges, respectively. Finally, Section~\ref{sec:ccl} discusses next priorities and concludes.

\section{Framing of the survey}
\label{sec:methodology}


\subsection{Objectives}

This survey focuses on NE recognition and classification, and does not consider entity linking nor entity relation extraction. With the overall objective of characterising the landscape of NER on historical documents, the survey reviews the history, the development, and the current state of related approaches. In particular, we attempt to answer the following questions:

\begin{enumerate}
    \item[\textbf{Q1}] What are the key challenges posed by historical documents to NER?
    \item[\textbf{Q2}] Which existing resources can be leveraged in this task, and what is their coverage in terms of historical periods, languages and domains?
    \item[\textbf{Q3}] Which strategies were developed and successfully applied in response to the challenges faced by NER on historical documents? Which aspects of NER systems require adaptation in order to obtain satisfying performances on this material? 
\end{enumerate}

\noindent While investigating the answers to these questions, the survey will also shed light on the variety of domains and usages of NE processing in the context of historical documents.

\subsection{Document Scope and Methodology}

Cultural heritage covers a wide range of material and the document scope of this survey, centred on `historical documents', needed to be clearly delineated. From a document processing perspective, there is no specific definition of what a historical document is, but only shared intuitions based on multiple criteria. Time seems an obvious one, but where to draw the line between historical and contemporary documents is a tricky question. 
Other aspects include the digital origin (digitised or born-digital), the type of writing (handwritten, typeset or printed), the state of the material (heavily degraded or not), and of the language (historical or not). 
None of these criteria define a clear set of documents 
and any attempt of 
definition resorts to, eventually, subjective decisions. 

In this survey, we consider as historical document any document of textual nature mainly, produced or published up to 1979, regardless of its topic, genre, style or acquisition method. 
The year 1979 is not arbitrary and corresponds to one of the most recent `turning points' acknowledged by historians~\cite{bosch_zeitenwende_2019}. This document scope is rather broad, and the question of the too far-reaching `textual nature' can be raised in relation to documents such as engravings, comics, card boards or even maps, which can also contain text. In practice, however, NER was mainly applied on printed documents so far, and these represent most of the material of the work reviewed here.

The compilation of the literature was based on the following strategies: scanning of the archives of relevant journals and conference series, search engine-based discovery, and  citation chaining. We considered key journals and conference series both in the fields of natural language processing and digital humanities (see Table \ref{tab:sources}). For searching, we used a combination of keywords over the Google Scholar and Semantic Scholar search engines.\footnote{E.g. `named entity recognition', `nerc', `named entity processing', `historical documents', `old documents' over \url{https://scholar.google.com} and \url{https://www.semanticscholar.org/}} With a few exceptions, we only considered publications that included a formal evaluation.

\ra{1.1}
\begin{table*}[t] 
\footnotesize
\centering
\begin{tabular}{lll}
\textbf{Title}  & \textbf{Type} & \textbf{Discipline} \\
\midrule
Annual Meeting of the Association for Computational Linguistics (ACL) & proceedings & CL/NLP \\
Digital Humanities conference & proceedings & DH \\
Digital Scholarship in the Humanities (DSH) & journal & DH \\
Empirical Methods in Natural Language Processing (EMNLP) & proceedings  & NLP \\
International Conference on Language Resources and Evaluation (LREC) & proceedings & NLP \\
International Journal on Digital Libraries & journal & DH \\
Journal of Data Mining and Digital Humanities (JDMDH) & journal & DH \\
Journal on Computing and Cultural Heritage (JOCCH) & journal & DH \\
Language Resources and Evaluation (LRE) & journal & NLP \\
SIGHUM Workshop on Computational Linguistics for Cultural Heritage & proceedings & CL/NLP/DH\\
\bottomrule
\end{tabular}
\caption{Publication venues whose archives were scanned as part of this survey (in alphabetical order).}
\label{tab:sources}
\vspace{-5mm}
\end{table*}

\subsection{Previous surveys and target audience}

Previous surveys on NER focused either on approaches in general, giving an overview of features, algorithms and applications, or on specific domains or languages. In the first group, Nadeau et al.~\cite{nadeau_survey_2007} provided the first comprehensive survey after a decade of work on NE processing, reviewing existing machine learning approaches of that time, as well as typologies and evaluation metrics. Their survey remained the main reference until the introduction of neural network-based systems, recently reviewed by Yadav et al.~\cite{yadav_survey_2018} and Li et al.~\cite{li_survey_2020}. The latest NER survey to date is the one by Nazar et al.~\cite{nasar_named_2021}, which focuses specifically on generic domains and on relation extraction. In the second group, Leaman et al.~\cite{leaman_banner_2008} and Campos et al.~\cite{campos_biomedical_2012} presented a survey of advances in biomedical named entity recognition, while Lei et al.~\cite{lei_research_2014} considered the same domain in Chinese. Shaalan focused on general NER in Arabic~\cite{shaalan_survey_2013}, and surveys exist for Indian languages~\cite{patil_survey_2016}. Recently, Georgescu et al.~\cite{georgescu_survey_2021} focused on NER aspects related to the cybersecurity domain. Turning our attention to digital humanities, Sporlerder~\cite{sporleder_natural_2010} and Piotrowski~\cite{piotrowski_natural_2012} provided general overviews of NLP processing for cultural heritage domains, considering institutional, documentary and technical aspects. To the best of our knowledge, this is the first survey on the application of NER to historical documents. 

The primary target audiences are researchers and practitioners in the fields of natural language processing and digital humanities, as well as humanities scholars interested in knowing and applying NER on historical documents. Since the focus is on adapting NER to historical documents and not on NER techniques themselves, this study assumes a basic knowledge of NER principles and techniques; however, it will provide information and guidance as needed. We use the terms `historical NER' and `modern NER' to refer to work and applications which focus on, respectively, historical and non-historical (as we define them) materials.

\section{Background}
\label{sec:background}

Before delving into NER for historical documents, this section provides a generic introduction to named entity processing and modern NER (Section \ref{sec:ner-general} and \ref{sec:ner-nutshell}), to the types of resources required (Section \ref{sec:ner-ressources}), and to the main principles underlying NER techniques (Section \ref{sec:ner-methods}).

\subsection{NE processing in general}
\label{sec:ner-general}

As of today, named entity tasks correspond to text processing steps of increasing level of complexity, defined as follows:
\begin{enumerate}
	\item recognition and classification -- or the detection of named entities, i.e. elements in texts which act as a rigid designator for a referent, and their categorisation according to a set of predefined semantic categories;
	\item disambiguation/linking -- or the linking of named entity mentions to a unique reference in a knowledge base, and 
	\item relation extraction -- or the discovery of relations between named entities.
\end{enumerate}

First introduced in 1995 during the 6$^{th}$ Message Understanding Conference~\cite{grishman_design_1995}, the task of NE recognition and classification (task 1 above) quickly broadened and became more complex, with the extension and refinement of typologies,\footnote{See e.g. 
the overviews of Nadeau et al.~\cite[pp. 3-4]{nadeau_survey_2007} and Ehrmann et al.~\cite{ehrmann_named_2016}.} the diversification of languages taken into account,  
and the expansion of the linguistic scope with, along proper names, the consideration of pronouns and nominal phrases as candidate lexical units (especially during the ACE program~\cite{doddington_automatic_2004}). Later on, as recognition and classification were reaching satisfying performances, attention shifted to finer-grained processing, with metonymy recognition \cite{markert_data_2009} and fine-grained classification~\cite{ekbal_assessing_2010,mai_empirical_2018}, and to the next logical step, namely entity resolution or disambiguation (task 2 above, not covered in this survey). Besides the general domain of clean and well-written news wire texts, NE processing is also applied to specific domains, particularly bio-medical~\cite{kim_genia_2003,goulart_systematic_2011}, and to more noisy inputs such as speech transcriptions~\cite{galibert_etape_2014} and tweets~\cite{ritter_named_2011,piskorski_named_2013}. In recent years, one of the major developments of NE processing  is its application to historical material.

Importantly, and although the question of the definition of named entities is not under focus here, we shall specify that we adopt in this regard the position of Nadeau et al.~\cite{nadeau_survey_2007} for which ``\textit{the word `Named' aims to restrict [Named Entities] to only those entities for which one or many rigid designators, as defined by S. Kripke, stands for the referent}''. Concretely speaking, named entities correspond to different types of lexical units, mostly proper names and definite descriptions, which, in a given discourse and application context, autonomously refer to a predefined set of entities of interest. 
There is no strict definition of named entities, but only a set of linguistic and application-related criteria which, eventually, compose a heterogeneous set of units.\footnote{See Ehrmann~\cite[pp.81-188]{ehrmann_named_2008} for an in-depth discussion of NE definition.}


Finally, let us mention two NE-related specific research directions: temporal information processing and geoparsing. This survey does not consider work related to temporal analysis and, when relevant, occasionally mentions some related to geotagging.

\subsection{NER in a nutshell}
\label{sec:ner-nutshell}

\subsubsection{A sequence labelling task}

Named entity recognition and classification is defined as a sequence labelling task where, given a sequence of tokens, a system seeks to assign labels (NE classes) to this sequence. The objective for a system is to observe, in a set of labelled examples, the word-labels correspondences and their most distinctive features in order to learn identification and classification patterns which can then be used to infer labels for new, unseen sequences of tokens. This excerpt from the CoNLL-03 English test dataset~\cite{tjongkimsang_introduction_2003} illustrates a training example (or the predictions a system should output):

\small

\begin{exe}
\ex{ $[_{LOC}$ Switzerland] stands accused by Senator $[_{PER}$ Alfonse D'Amato], chairman of the powerful $[_{ORG}$ U.S. Senate Banking Committee], of agreeing to give money to $[_{LOC}$ Poland] (...)}
\label{ex:conll03}
\end{exe}

\normalsize

\noindent Such input is often represented with the IOB tagging scheme, where each token is marked as being at the beginning (B), inside (I) or outside (O) of an entity of a certain class~\cite{ramshaw_text_1999}. 
Fig.~\ref{fig:iob} represents the above example in IOB format, from which systems try to extract features to learn NER models.

\subsubsection{Feature space}

NER systems' input corresponds to a linear representation of text as a sequence of characters, usually processed as a sequence of words and sentences.
This input is enriched with features or `clues' a system consumes in order to learn (or generalise) a model. Typical NER features may be observed at three levels: words, close context or sentences, and document. At the morphological level, features include e.g. the word itself, its length, whether it is (all) capitalised or not, whether it contains specific  word patterns or specific affixes (e.g. the suffixes \textit{-vitch} or \textit{-sson} for person names in Russian and Swedish), its base form, its part of speech (POS), and whether it is present in a predefined list. At the contextual level, features reflect the presence or absence of surrounding `trigger words' (or combination thereof, e.g. \textit{Senator} and \textit{to} preceding a person or location name, \textit{Committee} ending an organisation name), or of surrounding NE labels. Finally, at the document level, features correspond to e.g. the position of the mention in the document or paragraph, the occurrence of other entities in the document, or the document metadata. 

\ra{1.1}
\begin{table*}
\centering
\footnotesize
\caption{Illustration of IOB tagging scheme (example \ref{ex:conll03}).}
\begin{tabular}{llll}
\textbf{Tokens (X)}       & \textbf{NER label (Y)} & \textbf{POS} & \textbf{Chunk}  \\
\midrule
Switzerland & B-LOC & NNP & I-NP   \\
stands      & O     & VBZ & I-VP   \\
accused     & O     & VBN & I-VP   \\
by          & O     & IN  & I-PP   \\
Senator     & O     & NNP & I-NP   \\
Alfonse     & B-PER & NNP & I-NP   \\
D'Amato     & I-PER & NNP & I-NP   \\
...   & ... & ... & ...  
\end{tabular}
\label{fig:iob}
\end{table*}

These features can be absent or ambiguous, and none of them is systematically reliable; it is therefore necessary to combine them, and this is where statistical models are helpful. Features are observed in positive and negative examples, and are usually also encoded according to the IOB scheme (e.g. part-of-speech and chunk annotation columns in Fig.~\ref{fig:iob}). In traditional, feature-based machine learning, features are specified by the developer (feature engineering), while in deep learning they are learned by the system itself (feature learning) and go beyond those 
specified above.

\subsubsection{NER evaluation} Systems are evaluated in terms of precision (P), recall (R) and F-measure (F-score, the harmonic mean of P and R). Over the years, different scoring procedures and measures were defined in order to take into account various phenomena such as partial match or incorrect type but correct mention, or to assign different weights to various entity and/or error types. These fine-grained evaluation metrics allow for a better understanding of the system's performance and for tailoring the evaluation to what is relevant for an application. Examples include the (mostly abandoned) ACE `entity detection and recognition value' (EDR), the slot error rate (SER) or, increasingly, the exact vs. fuzzy match settings where entity mention boundaries need to correspond exactly vs. to overlap with the reference. We 
refer the reader to \cite[pp.12-15]{nadeau_survey_2007}, \cite[pp.3-4]{li_survey_2020} and \cite[chapter 6]{nouvel_entites_2015}. 
This survey reports systems' performances in terms of P, R and F-score.

\subsection{NER resource types}
\label{sec:ner-ressources}
Resources are essential when developing NER systems. Four main types of resources may be distinguished, each playing a specific role.

\subsubsection{Typologies} Typologies define a semantic framework for the entities under consideration. They corresponds to a formalised and structured description of the semantic categories to consider (the objects of the world which are of interest), along with a definition of their scope (their realisation in texts). There exist different typologies, which can be multi-purpose or domain-specific, and with various degrees of hierarchisation. Most of them are defined and published as part of evaluation campaigns, with no tradition of releasing typologies as such outside this context. Typologies form the basis of annotation guidelines, which explicit the rules to follow when manually annotating a corpus and are crucial for the quality of the resulting material.

\subsubsection{Lexicons and knowledge bases} Next, lexicons and knowledge bases provide information about named entities which may be used by systems for the purposes of recognition, classification and disambiguation. This type of resource has evolved significantly in the last decades, as a result of the increased complexity of NE-related tasks and of technological progress made in terms of knowledge representation. Information about named entities can be of lexical nature, relating to the textual units making up named entities, or of encyclop\ae{}dic nature, concerning their referents. The first case corresponds to simple lists named lexica or `gazetteers'\footnote{A term initially devoted to toponyms afterwards extended to any NE type.} which encode entity names, used in look-up procedures, and trigger words, used as features to guess names in texts. The second case corresponds to knowledge bases which encode various non-linguistic information about entities (e.g. date of birth/death, alma mater, title, function), used mainly for entity linking (Wikipedia
and DBpedia~\cite{lehmann_dbpedia_2015} being amongst the best-known examples). With the advent of neural language models, the use of explicit lexical information stored in lexica could have been definitely sealed, however gazetteer information still proves useful when incorporated as feature concatenated to pre-trained embeddings~\cite{collobert_natural_2011,huang_bidirectional_2015}, confirming that NER remains a knowledge-intensive task~\cite{ratinov_design_2009}.

\subsubsection{Word embeddings and language models}\label{sec:embedd} Word embeddings are low-dimensional, dense vectors which represent the meaning of words and are learned from word distribution in running texts. Stemming from the distributional hypothesis, they are part of the representation learning paradigm where the objective is to equip machine learning algorithms with generic and efficient data representations~\cite{bengio_representation_2013}. Their key advantage is that they can be learned in a self-supervised fashion, i.e. from unlabelled data, enabling the transition from feature engineering to feature learning. The principle of learning and using distributional word representations for different tasks was already present in \cite{baroni_distributional_2010,turian_word_2010,collobert_natural_2011}, but it is with the publication of word2vec, a software package which provided an efficient way to learn word embeddings from large corpora 
~\cite{mikolov_efficient_2013}, that embeddings started to become a standard component of modern NLP systems, including NER.

Since then, much effort has been devoted to developing effective means of learning word representations, first moving from words to sub-words and characters, and then from words to words-in-context with neural language models. The first generation of `traditional' embeddings corresponds to static word embeddings where a single representation is learned for each word independently of its context (at the type level). Common algorithms for such context-independent word embeddings include Google word2vec~\cite{mikolov_efficient_2013}, Stanford Glove~\cite{pennington_glove_2014} and SENNA~\cite{collobert_natural_2011}. The main drawbacks of such embeddings are their poor modelling of ambiguous words (embeddings are static) and their inability to handle out-of-vocabulary (OOV) words, i.e. words not present in the training corpus and for which there is no embedding. The usage of character-based word embeddings, i.e. word representations based on a combination of its character representations, can help process OOV words and make better use of morphological information. Such representations can be learned in a word2vec fashion, as with fastText~\cite{bojanowski_enriching_2017}, or via CNN or RNN-based architectures (see Section \ref{sec:ner-methods} for a presentation of types of networks).

However, even enriched with sub-word information, traditional embeddings are still ignorant of contextual information. This short-coming is addressed by a new generation of approaches which takes as learning objective language modelling, i.e. the task of computing the probability distribution of the next word (or character) given the sequence of previous words (or characters)~\cite{bengio_neural_2003}. By taking into account the entire input sequence, such approaches can learn deeper representations which capture many facets of language, including syntax and semantics, and are valid for various linguistic contexts (at the token level). They generate powerful language models (LMs) which can be used for downstream tasks and from which contextual embeddings can be derived. These LMs can be at the word level (e.g. ELMo~\cite{peters_deep_2018}, ULMFiT~\cite{howard_universal_2018}, BERT~\cite{devlin_bert_2019} and GPT~\cite{radford_improving_2018}), or character-based such as the contextual string embeddings proposed by Akbik et al.~\cite{akbik_contextual_2018} (a.k.a flair embeddings). Overall, alongside static character-based word and word embeddings, character-level and word-level LM embeddings are pushing the frontiers in NLP and are becoming key elements of NER systems, be it for contemporary or historical material. 

\subsubsection{Corpora} Finally, a last type of resource essential for developing NER systems is labelled documents and, to some extent, unlabelled textual data.
Labelled corpora illustrate an objective and are used either as a learning base 
or as a point of reference for evaluation purposes. Unlabelled textual material is necessary to acquire embeddings and language models.

\subsection{NER methods}
\label{sec:ner-methods} 

Similarly to other NLP tasks, NER systems are developed according to three standard families of algorithms, namely rule-based, feature-based (traditional machine learning) and neural-based (deep learning). 

\subsubsection{Rule-based approaches}
\label{sec:RB} 
Early NER methods in the mid-1990s were essentially rule-based. Such approaches rely on rules manually crafted by a developer (or linguist) on the base of regularities observed in the data. Rules manipulate language as a sequence of symbols and interpret associated information. Organised in what makes up a grammar, they often rely on a series of linguistic pre-processing (sentence splitting, tokenization, morpho-syntactic tagging), require external resources storing language information (e.g. triggers words in gazetteers) and are executed using transducers. 
Such systems have the advantages of not requiring training data and of being easily interpretable, but need time and expertise for their design. 

\subsubsection{Machine-learning based approaches}
\label{sec:ML} 
Very popular until the late 1990s, rule-based approaches were superseded by traditional machine learning approaches when large annotated corpora became available and allowed the machine learning of statistical models in supervised, semi-supervised, and later unsupervised fashion.
Traditional, feature-based machine learning algorithms learn inductively from data on the base of manually selected features. In supervised NER, they include support vector machines~\cite{isozaki_efficient_2002}, decision trees~\cite{szarvas_multilingual_2006}, as well as probabilistic sequence labelling approaches with generative models such as hidden markov models~\cite{bikel_nymble_1997} and discriminative ones such as maximum entropy models~\cite{bender_maximum_2003} 
and linear-chained conditional random fields (CRFs)~\cite{lafferty_conditional_2001}. 
Thanks to their capacity to take into account the neighbouring tokens, CRFs proved particularly well-suited for NER tagging and became the standard for feature-based NER systems. 

\subsubsection{Deep learning approaches}
\label{sec:DL} 
Finally, latest research on NER is largely (if not exclusively) dominated by deep learning (DL). 
Deep learning systems correspond to artificial neural networks with multiple processing layers which learn representations of data with multiple levels of abstraction~\cite{lecun_deep_2015}. In a nutshell, (deep) neural networks are composed of computational units, which take a vector of input values, multiply it by a weight vector, add a bias, apply a non-linear activation function, and produce a single output value. Such units are organised in layers which compose a network, where each layer receives its input from the previous one and passes it to the next (forward pass), and where parameters that minimise a loss function are learned with gradient descent (backward pass). 
The key advantage of neural networks is their capacity to automatically learn input representations instead of relying on manually elaborated features, and very deep networks (with many hidden layers) are extremely powerful in this regard.

Deep learning architectures for sequence labelling have undergone rapid change over the last few years. These developments are function of two decisive aspects for successful deep learning-based NER: at the architecture level, the capacity of a network to efficiently manage context, and, at the input representation level, the capacity to benefit from or learn powerful embeddings or language models. In what follows we briefly review main deep learning architectures for modern NER and refer the reader to Lin et al.~\cite{li_survey_2020} for more details.  

Motivated by the desire to avoid task-specific engineering as much as possible, Collobert et al.~\cite{collobert_natural_2011} pioneered the use of neural nets for four standard NLP tasks (including NER) with convolutional neural networks (CNN) that made used of trained type-level word embeddings and were learned in an end-to-end fashion. Their unified architecture SENNA\footnote{`Semantic/syntactic Extraction using a Neural Network Architecture'.} reached very competitive results for NER ($89.86\%$ F-score on the CoNLL-03 English corpus) and near state-of-the-art results for the other tasks. Following Collobert's work, developments focused on architectures capable of keeping information of the whole sequence throughout hidden layers instead of relying on fixed-length windows. These include recurrent neural networks (RNN), either simple~\cite{elman_finding_1990} or bi-directional~\cite{schuster_bidirectional_1997} (where input is processed from right to left and from left to right), and their more complex variants of long short-term memory networks (LSTM)~\cite{hochreiter_long_1997} and gated recurrent units (GRU)~\cite{cho_learning_2014} which mitigate the loss of distant information often observed in RNN. Huang et al.~\cite{huang_bidirectional_2015} were among the first to apply a bidirectional LSTM (BiLSTM) network with a CRF decoder to sequence labelling, obtaining $90.1\%$ F-score on the NER CoNLL-03 English dataset. Soon, BiLSTM networks became the de facto standard for context-dependent sequence labelling, giving rise to a body of work including Lample et al.~\cite{lample_neural_2016}, Chiu et al.~\cite{lample_neural_2016}, and Ma et al.~\cite{ma_endtoend_2016} 
(to name but a few). Besides making use of bidirectional variants of RNN, these work also experiment with various input representations, in most cases combining learned character-based representations with pre-trained word embeddings. Character information has proven useful for inferring information for unseen words and for learning morphological patterns, as demonstrated by the $91.2\%$ F-score of Ma et al.~\cite{ma_endtoend_2016} on CoNLL-03, and the systematically better results of Lample et al.~\cite{lample_neural_2016} on the same dataset when using character information. A more recent study by Taillé et al.~\cite{taille_contextualized_2020} confirms the role of sub-word representations for unseen entities.

The latest far-reaching innovation in the DL architecture menagerie corresponds to self-attention networks, or transformers~\cite{vaswani_attention_2017}, a new type of simple networks which eliminates recurrence and convolutions and are based solely on the attention mechanism. Transformers allow for keeping a kind of global memory of the previous hidden states where the model can choose what to retrieve from (attention), and therefore use relevant information from large contexts. They are mostly trained with a language modelling objective and are typically organised in transformer blocks, 
which can be stacked and used as encoders and decoders. 
Major pre-training transformer architectures include the Generative Pre-trained Transformer (GPT, a left-to-right architecture)~\cite{radford_improving_2018} and the Bidirectional Encoder Representation from Transformer (BERT, a bidirectional architecture)~\cite{devlin_bert_2019}, which achieves $92.8\%$ NER F-score on CoNLL-03. More recently, Yamada et al.~\cite{yamada_luke_2020} proposed an entity-aware self-attention architecture 
which achieved $94.3\%$ F-score on the same dataset. Transformer-based architectures are the focus of extensive research and many model variants were proposed, of which Tay et al.~\cite{tay_efficient_2020} propose an overview.

Overall, two points should be noted. First, that beyond the race for the leader board (based on the fairly clean 
English CoNLL-03 dataset), pre-trained embeddings and language models play a crucial role and are becoming a new paradigm in neural NLP and NER (the `NLP's ImageNet moment'~\cite{ruder_nlp_2018}). Second, that powerful language models are also paving the way for transfer learning, a method particularly useful with low-resource languages and out-of-domain contexts, as is the case with challenging, historical texts.

\section{Challenges}
\label{sec:challenges}

Named entity recognition on historical documents faces four main challenges for which systems developed on contemporary datasets are often ill-equipped. Those challenges are intrinsic to the historical setting, like time evolution and types of documents, and endemic to the text acquisition process, like OCR noise. This translates into a variable and sparse feature space, a situation compounded by the lack of resources. This section successively considers the challenges of document type and domain variety, noisy input, dynamics of language, and lack of resources.

\subsection{The (historical) variety space}

First, NER on historical texts corresponds to a wide variety of settings, with documents of different types (e.g. administrative documents, media archives, literary works, documentation of archival sites or art collections, correspondences, secondary literature), of different nature (e.g. articles, letters, declarations, memoirs, wires, reports), and in different languages, which, moreover, spans different time periods and encompasses various domains and countless topics. The objective here is not to inventory all historical document types, domains and topics, but to underline the sheer variety of settings which, borrowing an expression from B. Plank~\cite{plank_what_2016}, compose the `variety space' NLP is confronted with, intensified in the present case by the time dimension.\footnote{Considering there is no common grounds on what constitutes a domain and that the term is overloaded, 
Plank proposes the concept of ``variety space'', defined as a ``\textit{unknown high-dimensional space, whose dimensions contain (fuzzy) aspects such as language (or dialect), topic or genre, and social factors (age, gender, personality, etc.), amongst others. A domain forms a region in this space, with some members more prototypical than others}''~\cite{plank_what_2016}.}

Two comments should be made in connection with this variety. First, domain shift is a well-known issue for NLP systems in general and for modern NER in particular. While B. Plank~\cite{plank_what_2016} and J. Einsenstein~\cite{eisenstein_what_2013} investigated what to do about bad and non-standard (or non-canonical) language with NLP in general, Augenstein et al.~\cite{augenstein_generalisation_2017} studied the ability of modern NER systems to generalise over a variety of genres, and Taillé et al.~\cite{taille_contextualized_2020} over unseen mentions. Both studies demonstrated a NER transfer gap between different text sources and domains, confirming earlier findings of Vilain et al.~\cite{vilain_entity_2007}. While no studies have (yet) been conducted on the generalisation capacities of NER systems within the realm of historical documents, there are strong grounds to believe that systems are equally impacted when switching domain and/or document type.

Second, this (historical) variety space is all the more challenging as the scope of needs and applications in humanities research is much broader than the one usually addressed in modern NLP. For sure the variety space does not differ much between today and yesterday's documents (i.e. if we were NLP developers living in the 18C we would be more or less confronted with the same `amount' of variety as today), however here the difference lies in the interest for all or part of this variety: while NLP developments tend to focus on some well-identified and stable domains/sub-domains (sometimes motivated by commercial opportunities), the (digital) humanities and social sciences research communities are likely interested in the whole spectrum of document types and domains. In brief, if the magnitude of the variety space is more or less similar for contemporary and historical documents, the range of interests and applications in humanities and cultural heritage requires---almost by design---the consideration of an expansive array of domains and document types.

\subsection{Noisy input}

Next, historical NER faces the challenges of noisy input derived from automatic text acquisition over document facsimiles. Text is acquired via two processes: 1) optical character recognition (OCR) and handwritten text recognition (HTR), which recognise text characters from images of printed and handwritten documents respectively, and 2) optical layout recognition (OLR), which identifies, orders and classifies text regions (e.g. paragraph, column, header). We consider both successively.

\subsubsection{Character recognition}

The OCR transcription of the newspaper article on the right-hand side of Figure \ref{fig:newspaper} illustrates a typical, mid-level noise, with words perfectly readable (\textit{la Belgique}), others illegible (\textit{pu. s >s « \_jnces}), and tokenization problems (\textit{n'à'pas}, \textit{le'Conseiller}). While this does not really affect human understanding when reading, the same is not true for machines which face numerous OOV words. Be it by means of OCR or HTR, text acquisition performances can be impacted by several factors, including: a) the quality of the material itself, affected by the poor preservation and/or original state of documents with e.g. ink bleed-through, stains, faint text, and paper deterioration; b) the quality of the scanning process, with e.g. an inadequate resolution or imaging process leading to frame or border noise, skew, blur and orientation problems; 
or c) as per printed documents and in absence of standardisation, the diversity of typographic conventions through time including e.g. varying fonts, mixed alphabets but also diverse shorthand, accents and punctuation. 
These difficulties naturally challenge character recognition algorithms which are, what is more, evolving from one OCR campaign to another, usually conducted at different times by libraries and archives. As a result, not only the transcription quality is below expectations, but the type of noise present in historical machine-readable corpora is also very heterogeneous. 

Several studies investigated the impact of OCR noise on downstream NLP tasks. While Lopresti~\cite{lopresti_optical_2009} demonstrated the detrimental effect of OCR noise propagation through a typical NLP pipeline on contemporary texts, Van Strien et al.~\cite{vanstrien_assessing_2020} focused on historical material and found a consistent impact of OCR noise on the six NLP tasks they evaluated. If sentence segmentation and dependency parsing bear the brunt of low OCR quality, NER is 
also affected with a significant drop of F-score between good and poor OCR (from $87\%$ to $63\%$ for person entities). Focusing specifically on entity processing, Hamdi et al.~\cite{hamdi_analysis_2019,hamdi_assessing_2020} confronted a BiLSTM-based NER model with OCR outputs of the same text but of different qualities and observed a 30 percentage point loss in F-score when the character error rate increased from 7\% to 20\%. 
Finally, in order to assess the impact of noisy entities on NER during the CLEF-HIPE-2020 NE evaluation campaign on historical newspapers (HIPE-2020 for short),\footnote{\url{https://impresso.github.io/CLEF-HIPE-2020/}} Ehrmann et al.~\cite{ehrmann_extended_2020} evaluated systems' performances on various entity noise levels, defined as the length-normalised Levenshtein distance between the OCR surface form of an entity and its manual transcription. They found remarkable performance differences between noisy and non-noisy mentions, and that already as little noise as 0.1 severely hurts systems' abilities to predict an entity and may halve their performances. To sum up, whether focused on a single OCR version of text(s)~\cite{vanstrien_assessing_2020}, on different artificially-generated ones~\cite{hamdi_analysis_2019}, or on the noise present in entities themselves~\cite{ehrmann_extended_2020}, these studies clearly demonstrate how challenging OCR noise is for NER systems.

\subsubsection{Layout recognition}

Beside incorrect character recognition, textual input quality can also be affected by faulty layout recognition. Two problems surface here. The first relates to incorrect page region segmentation which mixes up text segments and produces, even with correct OCR, totally unsuitable input (e.g. a text line reading across several columns). Progress in OLR algorithms makes this problem rarer, but it is still present for collections processed more than a decade ago. The second has to do with the unusual text segmentation resulting from correct OLR of column-based documents, with very short line segments resulting in numerous hyphenated words (cf. Figure \ref{fig:newspaper}). The absence of proper sentence segmentation and word tokenization also affects performances, as demonstrated in HIPE-2020
, in particular Boros et al~\cite{boros_robust_2020}, \citet{ortizsuarez_sinner_2020} and Todorov et al.~\cite{todorov_transfer_2020} (see Section \ref{dla}).

\vspace{0.15cm}
Overall, OCR and OLR noises lead to a sparser feature space which greatly affects NER performances. What makes this `noisiness' particularly challenging is its wide diversity and range: an input can be noisy in many different ways, and be little to very noisy. Compared to social media, for which Baldwin et al.~\cite{baldwin_how_2013} demonstrated that there exists a noise similarity from a medium to another (blog, Twitter, etc.) and that this noise is mostly `NLP-tractable', OCR and OLR noises in historical documents appear as real moving targets.

\subsection{Dynamics of language}

Another challenge relates to the effects of time and the dynamics of language. As a matter of fact, historical languages exhibit a number of differences with modern ones, having an impact on the performances of NLP tools in general, and of NER in particular~\cite{piotrowski_natural_2012}.

\subsubsection{Historical spelling variations} The first source of difficulty relates to spelling variations across time, due either to the normal course of language evolution or to more prescriptive orthographic reforms. For instance, the 1740 edition of the dictionary of the French Academy (which had 8 editions between 1694 and 1935) introduced changes in the spelling of about one third of the French vocabulary and, in Swedish 19C literary texts, the letters <f/w/e/q> were systematically used instead of <v/v/ä/k> in modern Swedish~\cite{borin_naming_2007}. NER can therefore be affected by poor morpho-syntactic tagging over such morphological variety, and by spelling variation of trigger words and of proper names themselves. While the latter are less affected by orthographic reforms, they do vary through time~\cite{borin_naming_2007}.

\subsubsection{Naming conventions} Changes in naming conventions, particularly for person names, can also be challenging. 
Let alone the numerous aristocratic and military titles that were used in people's addresses, it was, until recently, quite common to refer to a spouse using the name of her husband (which affects more the linking than recognition), and to use now outdated addresses, e.g. the French 
expression \textit{sieur}. These changes have been studied by Rosset et al.~\cite{rosset_structured_2012} who compared the structure of entity names in historical newspapers vs. in contemporary broadcast news. Differences include the prevalence of the structure \textit{title + last name} vs. \textit{first + last name} for \textit{Person} in historical newspapers and contemporary broadcast news respectively, and of single-component names vs. multiple-component names for \textit{Organisation} (idem). Testing several classifiers, the authors also showed that it is possible to predict the period of a document from the structure of its entities, thus confirming the evolution of names over time. For their part, Lin et al.~\cite{lin_rigorous_2020} studied the generalisation capacities of a state-of-the-art neural NER system on entities with weak name regularity in a modern corpus 
and concluded that name regularity is critical for supervised NER models to generalise over unseen mentions.

\subsubsection{Entity and context drifts} Finally, a further complication comes from the historicity of entities, also known as entity drift, with places, professions, and types of major entities fading and emerging over time. For instance, a large part of profession names, which can be used as clues to recognise persons, has changed from the 19C to the 21C.\footnote{See for example the variety of occupations in the HISCO database: \href{https://iisg.amsterdam/en/data/data-websites/history-of-work}{iisg.amsterdam/en/data/data-websites/history-of-work}} 
This dynamism is still valid today (NEs are an open class) and its characteristics as well as its impact on performances is particularly well documented for social media: Fromreide et al. showed a loss of 10 F-score percentage points between two Twitter corpora sampled two years apart~\cite{fromreide_crowdsourcing_2014}, and Derczynski et al. systematised the analysis with the W-NUT2017 shared task on novel and emerging entities where, on training and test sets with very little entity overlaps, the maximum F-score was only $40\%$~\cite{derczynski_results_2017}. Besides confirming some degree of `artificiality' of classical NE corpora where the overlap between mentions in the train and the test sets do not reflect real-life settings, these studies illustrate the poor generalisation capacities of NER systems to unseen mentions due to time evolution. How big and how quick is entity drift in historical corpora? We could not find any quantitative study on this, but a high variability of the global referential frame through time is more than likely.

\vspace{0.15cm}
Overall, the dynamics of language represent a multi-faceted challenge where the disturbing factor is not anymore an artificially introduced noise like with OCR and OLR, but the naturally occurring alteration of the signal by the effects of time. Both phenomena result in a sparser feature space, but the dynamics of language appear less elusive and volatile than OCR. 
Compared to OCR noise, its impact on NER performances is however relatively under-studied, and only a few diachronic evaluations were conducted on historical documents so far. Worth of mention is the evaluation of several NER systems on historical newspaper corpora spanning ca. 200 years, first with the study of Ehrmann et al.~\cite{ehrmann_diachronic_2016}, second on the occasion of the HIPE-2020 shared task~\cite{ehrmann_extended_2020}. Testing the hypothesis of the older the document, the lower the performance, both studies reveal a contrasted picture with non-linear F-score variations over time. If a clear trend of increasing recall over time can be observed in \cite{ehrmann_diachronic_2016}, further research is needed to distinguish and assess the impact of each of the aforementioned time-related variations.

\subsection{Lack of resources}
\label{sec:lack-resources}

Finally, the three previous challenges are compounded by a fourth one, namely a severe lack of resources. As mentioned in Section~\ref{sec:ner-ressources}, the development of NER systems relies on four types of resources---typologies, lexicons, embeddings and corpora---which are of particular importance for the adaptation of NER systems to historical documents.

With respect to typologies, the issue at stake is, not surprisingly, their dependence on time and domain. While mainstream typologies with few `universal' classes (e.g. \textit{Person}, \textit{Organisation}, \textit{Location}, and a few others) can for sure be re-used for historical documents, this obviously does not mean that they are perfectly suited to the content or application needs of any particular historical collection. Just as universal entity types cannot be used in all contemporary application contexts
, neither can they be systematically applied to all historical documents: only a small part can be reused, and they require adaptation. An example is warships, often mentioned in 19C documents, for which none of the mainstream typologies has an adequate class. To say that typologies need to be adapted is almost a truism, but it is worth mentioning for it implies that the application of off-the-shelf NER tools--as is often done--is unlikely to capture all entities of interest in a specific collection and, therefore, is likely to penalise subsequent studies.  

Besides the (partial) inadequacy of typologies, the lack of annotated corpora severely impedes the development of NER systems for historical documents, for both training and evaluation purposes. While unsupervised domain adaptation approaches are gaining interest~\cite{ramponi_neural_2020}, most methods still depend on labelled data to train their models. Little training data usually results in inferior performances, as demonstrated---if proof were needed---by Augenstein et al. for NER on contemporary data~\cite[p. 71]{augenstein_generalisation_2017}, and by Ehrmann et al. on historical newspapers~\cite[Section 7]{ehrmann_extended_2020}. NE-annotated historical corpora exist, but are still rare and scattered over time and domains (cf. Section \ref{sec:resources}). This paucity also affects systems' evaluation and comparison which, besides the lack of gold standards, is also characterised by fragmented and non-standardised evaluation approaches. The recently organised CLEF-HIPE-2020 shared task on NE processing in multilingual and historical newspapers is a first step towards alleviating this situation~\cite{ehrmann_extended_2020}.

Last but not least, if large quantities of textual data are being produced via digitisation, several factors slow down their dissemination and usage as base material to acquire embeddings and language models. First, textual data is acquired via a myriad of OCR softwares which, despite the definition of standards by libraries and archives, supply quite disparate and heavy-to-process output formats~\cite{ehrmann_language_2020,romanello_impresso_2020}. Second, even when digitised, historical collections are not systematically openly accessible due to copyright restrictions. 
Despite the recent efforts and the growing awareness of cultural institutions of the value of such assets for machine learning purposes~\cite{padilla_responsible_2020}
, these factors still hamper the learning of language representations from large amounts of historical texts.

Far from being unique to historical NER, lack of resources is a well-known problem in modern NER~\cite{ehrmann_named_2016}, and more generally in NLP~\cite{joshi_state_2020}. In the case at hand, the lack of resources is exacerbated by the somewhat youth of the research field and the relatively low attention towards the creation of resources compared to other domains. Moreover, considering how wide is the spectrum of domains, languages, document types and time periods to cover, it is likely that a certain resource sparsity will always remain. Finding ways to mitigate the impact of the lack of resources on system development and performances is thus essential.

\vspace{0.15cm}
\noindent \textbf{Conclusion on challenges}. NER on historical documents faces four main challenges, namely historical variety space, noisy input, dynamics of language, and lack of resources. If none of these challenges is new per se---which does not lessen their difficulty---, what makes the situation particularly challenging is their combination, in what could somehow be qualified an `explosive cocktail'. This set of challenges has two main characteristics: first, the prevalence of the time dimension, which not only affects language and OCR quality but also causes domain and entity drifts; and, second, the intensity of the present difficulties, with OCR noise being a real moving target, and domains and (historical) languages being highly heterogeneous. As a result, with feature sparsity adding up to multiple confounding factors, systems' learning capacities are severely affected. 
NER on historical documents can therefore be cast as a domain and time adaptation problem, where approaches should be robust to non-standard, historical inputs, what is more in a low-resource setting. A first step towards addressing these challenges is to rely on appropriate resources, discussed in the next section.

\section{Resources for Historical NER}
\label{sec:resources}

This section surveys existing resources for historical NER, considering typologies and annotation guidelines, annotated corpora, and language representations (see Section \ref{sec:ner-ressources} for a presentation of NER resource types). Special attention is devoted to how these resources distribute over languages, domains and time periods, in order to highlight gaps that future efforts should attempt to fill.

\subsection{Typologies and annotation guidelines}

Typologies and annotation guidelines for modern NER cover primarily the general and bio-medical domains, and the most used ones such as MUC~\cite{grishman_message_1996}, CoNLL~\cite{tjongkimsang_introduction_2003}, and ACE~\cite{doddington_automatic_2004} consist mainly of a few high-level classes with the `universal' triad \textit{Person}, \textit{Organisation} and  \textit{Location}~\cite{ehrmann_named_2016}. 
Although they are used in various contexts, they do not necessarily cover the needs of historical documents.
To the best of our knowledge, very few typologies and guidelines designed for historical material were publicly released so far. Exceptions include the Quaero~\cite{rosset_entites_2011,rosset_structured_2012}, SoNAR~\cite{menzel_guidelines_2021} and \textit{impresso} (used in HIPE-2020)~\cite{ehrmann_impresso_2020} typologies and guidelines adapted or developed for historical newspapers in French, German, and English. 
Designing guidelines and effectively annotating NEs in historical documents is not as easy as it sounds and  peculiarities of historical texts must be taken into account. These include for example OCRed text, with the question of how to determine the boundaries of mentions in gibberish strings, and historical entities, with the existence of various historical statuses of entities through times (e.g. \textit{Germany} has 8 Wikidata IDs over the 19C and 20C \cite[pp.9-10]{ehrmann_overview_2020}).

\subsection{Annotated corpora}

Annotated corpora correspond to sets of documents manually or semi-automatically tagged with NEs according to a given typology, and are essential for the development and evaluation of NER systems (see Section~\ref{sec:ner-ressources}). This section inventories NE-annotated historical corpora documented in publications and released under an open license.\footnote{Inventory as of June 2021. The \textit{Voices of the Great War} corpus~\cite{boschetti_voices_2020} is not included for not released under an open license.} Their presentation is organised into three broad groups (`news', `literature(s)' and `other'), where they appear in alphabetical order. Unless otherwise noted, all corpora consist of OCRed documents.

Let us start with some observations on the general picture. We could inventory 17 corpora, whose salient characteristics are summarised in Table \ref{tab:corpora}. It is worth noting that collecting information about released corpora is far from easy and that our descriptions are therefore not homogeneous. In terms of language coverage, the majority of corpora are monolingual, and less than a third include documents written in two or more languages. Overall, these corpora provide support for eleven currently spoken languages and two dead languages (Coptic and Latin). With respect to corpus size, the number of entities appears as the main proxy and we distinguish between small (< 10k), medium (10-30k), large (30-100k) and very large corpora (> 100k).\footnote{For comparison, the CoNLL-03 dataset contains ca. 70k mentions for English and 20k for German~\cite{tjongkimsang_introduction_2003}, while OntoNotes v5.0 contains 194k mentions for English, 130k for Chinese and 34k for Arabic~\cite{pradhan_conll2012_2012}.} In the present inventory, very large corpora are rather exceptional; roughly one third of them are small-sized, while the remaining are medium- or large-sized corpora. Next, and not surprisingly, a wide spectrum of domains is represented, from news to literature. This tendency towards domain specialisation is also reflected in typologies with, alongside the ubiquitous triad of \textit{Person}, \textit{Location}, and \textit{Organisation} types, a long tail of specific types reflecting the information or application needs of particular domains. Finally, in terms of time periods covered, we observe a high concentration of corpora in the 19C, directly followed by 20C and 21C, while corpora for previous centuries are either scarce or absent.

\subsubsection{News}
\label{sec:news}
The first group brings together corpora built from historical newspaper collections. With corpora in five languages (Czech, Dutch, English, French and German), news emerges as the best-equipped domain in terms of labelled data availability.

The Czech Historical NE Corpus
~\cite{hubkova_czech_2020} is a small corpus produced out of the year 1872 of the Czech title \textit{Posel od Čerchova}. Articles are annotated according to six entity types---persons, institutions, artifacts \& objects, geographical names, time expressions and ambiguous entities---which, despite being custom, bear substantial similarities with major typologies. The corpus was manually annotated by two annotators with an inter-annotator agreement (IAA) of 0.86 (Cohen's Kappa).

Europeana NER corpora\footnote{\url{https://github.com/EuropeanaNewspapers/ner-corpora}}~\cite{neudecker_open_2016} is a large-sized collection of NE-annotated historical newspaper articles in Dutch, French and German, containing primarily 19C materials. These corpora were sampled from the Europeana newspaper collection~\cite{neudecker_making_2016} 
by randomly selecting 100 pages from all titles for each language, considering only pages with a minimum word-level accuracy of 80\%. 
Three entity types were considered (person, location, organisation), yet no IAA for the annotations is reported. Instead, the quality and usefulness of these annotated corpora were assessed by training and evaluating the Stanford CRF NER classifier (see Section \ref{sec:ML}).

\ra{1.3}
\begin{table*}[t]
\scriptsize
\centering
\begin{tabular}{llllllrl}

\textbf{Corpus}  & \textbf{Doc. type} & \textbf{Time period} & \textbf{Tag set} & \textbf{Lang.} & \textbf{\# NE}s & \textbf{Size} & \textbf{License} \\
\midrule
Quaero Old Press~\cite{rosset_structured_2012}         & newspapers     & 19C     & Quaero  & fr & 147,682 & \textsc{xl} & \textsc{elra}\\
Europeana~\cite{neudecker_open_2016}                   & newspapers     & 19C     & \textsc{per,loc,org} & fr, de, nl & 40,801 & \textsc{l} &\textsc{cc0}  \\
De Gasperi~\cite{sprugnoli_fifty_2016}          & various types  & 20C & \textsc{per}, \textsc{gpe} & it & 35,491 & \textsc{l} &\textsc{cc by-nc-sa} \\
Latin NER~\cite{erdmann_challenges_2016}        & literary texts & 1C~\textsc{bce}-2C & \textsc{per,geo,grp} & la & 7,175 & \textsc{s} &\textsc{gpl v3.0}\\
HIMERA~\cite{thompson_text_2016}                & medical lit.   & 19C-21C & custom   & en    & 8,400    & \textsc{s} &\textsc{cc by} \\
Venetian references~\cite{colavizza_annotated_2017}    & publications   & 19C-21C  & custom & Multi & 12,879 & \textsc{m} &\textsc{cc by} \\
Finnish NER~\cite{ruokolainen_recherche_2018}   & newspapers     & 19C-20C  & \textsc{per,loc,org} & fi & 26,588 & \textsc{m} & n/a  \\
\textsc{droc}~\cite{krug_description_2018}    & novels         & 17C-20C  & custom & de & 6,013 &  \textsc{s} &\textsc{cc by} \\
Travel writings~\cite{sprugnoli_arretium_2018}         & travelogues    & 19C-20C  & \textsc{loc} & en & 2,228 & \textsc{s} & n/a \\
Czech Hist. NE Corpus~\cite{hubkova_namedentity_2019}  & newspapers     & 19C      & custom               & cz  & 4,017 & \textsc{s} &\textsc{\textsc{cc by-nc-sa}} \\
LitBank~\cite{bamman_annotated_2019}                   & novels         & 19C-20C  & \textsc{ace} (w/o \textsc{wea}) & en & 14,000 & \textsc{l} &\textsc{cc by-sa} \\
BIOfid~\cite{ahmed_biofid_2019}                & publications   & 18C-20C  & extended GermEval  & de & 33,545 & \textsc{l} &\textsc{gpl v3.0} \\
HIPE~\cite{ehrmann_overview_2020}               & newspapers     & 18C-21C  & \textit{impresso}    & de, en, fr & 19,848 & \textsc{m} &\textsc{cc by-nc-sa} \\
BDCamões~\cite{grilo_bdcamoes_2020}                    & literary texts & 16C-21C  & custom & pt & 144,600 & \textsc{xl} &\textsc{cc by-nc-nd} \\
Coptic Scriptorium corpora   & literary texts & 3C-5C    & custom & cop & 88,068 & \textsc{l} &\textsc{cc by} \\
GeoNER~\cite{kogkitsidou_normalisation_2020}              & literary texts & 16C-17C  & \textsc{geo} & fr & 264 & \textsc{s} &\textsc{lgpl-lr} \\
NewsEye~\cite{hamdi_multilingual_2021}          & newspapers     & 19C-20C & \textit{impresso}-comp. & de, fr, fi,s v & 30,580 & \textsc{l} &\textsc{cc by} \\

\bottomrule
\end{tabular}
\caption{Overview of reviewed NE-annotated historical corpora (ordered by publication year).}
\label{tab:corpora}
\vspace{-4mm}
\end{table*}

The Finnish NER corpus\footnote{\url{https://digi.kansalliskirjasto.fi/opendata/submit} (Digitalia (2017-2019) package).}
~\cite{ruokolainen_recherche_2018} is composed of a selection of pages from journals and newspapers published between 1836 and 1918 and digitized by the national library of Finland. The OCR of this medium-size corpus was manually corrected by librarians and NE annotations were made manually for half of them, semi-automatically for the other (via the manual correction of the output of a Stanford NER system trained on the manually corrected subset). Overall, the annotations show a good IAA of 0.8 (Cohen's kappa).

The HIPE corpus\footnote{Version 1.3, \url{https://github.com/impresso/CLEF-HIPE-2020/tree/master/data}}~\cite{ehrmann_overview_2020} is a medium-sized, historical news corpus in French, German and English, created as part of HIPE-2020. It consists of newspaper articles sampled from Swiss, Luxembourgish and American newspaper collections covering a time span of ca. 200 years (1798-2018). 
OCR quality of the corpus corresponds to real-life setting and varies depending on the digitisation time and preservation state of original documents.
The corpus was annotated following the \textit{impresso} guidelines~\cite{ehrmann_impresso_2020}, which are based on and are retro-compatible with the Quaero guidelines~\cite{rosset_entites_2011}. The annotation tag set comprises 5 coarse-grained and 23 fine-grained entity types, and includes entity components as well as nested entities. Wrongly OCRed entity surface forms are manually corrected and entities are linked towards Wikidata. NERC and EL annotations reached an average IAA across languages of 0.8 (Krippendorf's alpha).

The NewsEye dataset\footnote{Version 1.0, \url{https://doi.org/10.5281/zenodo.4573313}} \cite{hamdi_multilingual_2021} is a large-sized corpus composed of articles extracted from newspapers published between mid 19C and mid 20C in French, German, Finnish, and Swedish. Four entity types were considered (person, location, organisation and human product) and annotated according to guidelines\footnote{\url{https://zenodo.org/record/4574199}} similar to the \textit{impresso} ones; entities are linked towards Wikidata and articles are further annotated with authors' stances. The annotation reaches high IAAs exceeding 0.8 for Swedish and 0.9 for German, French and Swedish (Cohen’s kappa). 

The Quaero Old Press Extended NE corpus\footnote{\url{http://catalog.elra.info/en-us/repository/browse/ELRA-W0073/}}~\cite{rosset_structured_2012} is a very large annotated corpus composed of 295 pages sampled from French newspapers 
of December 1890. 
The OCR quality is rather good, with a character and word error rates of 5\% and 36.5\% respectively. Annotators were asked to transcribe wrongly OCRed entity surface forms---similarly to what was done for the HIPE corpus---which makes both corpora suitable to check the robustness of NER systems to OCR noise. The annotator agreement on this corpus reaches 0.82 (Cohen's Kappa).

\subsubsection{Literature(s)}
\label{sec:lit}

The second group of corpora relates to literature and is more heterogeneous in terms of domains and document types, ranging from literary texts to scholarly publications.

To begin with, two resources consist of ancient literary texts. 
First, the Latin NER corpus\footnote{\url{https://github.com/alexerdmann/Herodotos-Project-Latin-NER-Tagger-Annotation}}~\cite{erdmann_challenges_2016} comprises ancient literary material sampled from three texts representatives of different literary genres (prose, letters and elegiac poetry) and spanning over three centuries. The annotation tag set covers persons, geographical place names and group names (e.g. `Haeduos', a Gallic tribe). 
Next, the Coptic Scriptorium corpus\footnote{\url{https://github.com/copticscriptorium/corpora}} is a large-sized collection of literary works written in Coptic, the language of Hellenistic era Egypt (3C-5C CE), and belonging to multiple genres (hagiographic texts, letters, sermons, martyrdoms and the Bible). Besides lemma and POS tags, this corpus also contains (named and non-named) entity annotations, with links towards Wikipedia.
In addition to persons, places and organisations, the entity types include abstract entities (e.g. `humility'), animals, events, objects (e.g. `bottles'), substances (e.g. `water') and time expressions. Entity annotations were produced automatically (resulting in 11k named entities and 6k linked entities), a subset of which was manually corrected (2,4k named entities and 1,5k linked entities). 

Then, several corpora were designed to support computational literary analysis.
This is the case of the BDCamões Collection of Portuguese Literary Documents\footnote{\url{https://portulanclarin.net/}}~\cite{grilo_bdcamoes_2020}, a very large annotated corpus composed of 208 OCRized texts (4 million words) representative of 14 literary genres and covering five centuries of Portuguese literature (16C-21C). 
Due to the large time span covered, texts adhere to different orthographic conventions. 
Named entity annotations correspond to locations, organisations, works, events and miscellaneous entities, and were automatically produced (silver annotations). They constitute only one of the many layers of linguistic annotations of this corpus, alongside POS tags, syntactic analysis and semantic roles. 
Next, the LitBank\footnote{\url{https://github.com/dbamman/litbank}}~\cite{bamman_annotated_2019} dataset is a medium-sized corpus composed of 100 English literary texts published between mid 19C and beginning 20C.
Entities were annotated following the ACE guidelines---with the only exception of weapons as rarely attested---and include noun phrases as well as nested entities. 
Finally, the Deutsches ROman Corpus (DROC)~\cite{krug_description_2018} is a set of 90 richly-annotated fragments of German novels published between 1650 and 1950.
The DROC corpus is enriched with character mentions, character co-references, and direct speech occurrences. 
It features more than 50,000 character mentions, of which only 12\% (6,013) contain proper names and thus correspond to traditional person entity mentions (others correspond to pronouns or appellatives). 

Next, two of the surveyed corpora in this group focus specifically on place names. First, Travel writings\footnote{\url{https://github.com/dhfbk/Detection-of-place-names-in-historical-travel-writings}}~\cite{sprugnoli_arretium_2018} is a small corpus of 38 English travelogues printed between 1850 and 1940. Its tag set consists of a single type (\textit{Location}), which encompasses geographical, political and functional locations, thus corresponding to ACE's \textsc{gpe}, \textsc{loc} and \textsc{fac} entity types altogether. Second, the GeoNER corpus\footnote{\url{https://github.com/PhilippeGambette/GeoNER-corpus}}\cite{kogkitsidou_normalisation_2020} is a very small corpus consisting of three 16C-17C French literary texts by Racine, Molière and Marguerite de Valois. Each annotated text is available in its original version, as well as with automatic and manual historical spelling normalization. Despite its limited size, this corpus can be a valuable resource for researchers investigating the effects of historical normalisation on NER. 

Finally, moving from literature to scholarly literature, three corpora should be mentioned. First, BIOfid\footnote{\url{https://github.com/FID-Biodiversity/BIOfid}}~\cite{ahmed_biofid_2019} is a large NE-annotated corpus composed of ca. 1000 articles sampled from German books and scholarly journals in the domain of biodiversity and published between 18C and 20C. 
The annotation guidelines used for this corpus build upon those used for the GermEval dataset~\cite{benikova_nostad_2014}, with the addition of 
time expressions 
and taxonomies (\textit{Taxon}), i.e. systematic classifications of organisms by their characteristics (e.g. ``northern giant mouse lemur'').
Second, HIstory of Medicine CoRpus Annotation (HIMERA)\footnote{\url{http://www.nactem.ac.uk/himera/}}~\cite{thompson_text_2016} is a small-sized  corpus in the domain of medical history, consisting of journal articles and medical reports published between 1840 and 2013. 
This corpus is annotated with NEs according to a custom typology comprising, for example, medical conditions, symptoms, or biological entities. 
While all annotations were performed on manually corrected OCR output, the annotation of certain types was carried out in a semi-automatic fashion. Globally, the annotation reaches good IAAs of 0.8 and 0.86 for exact and relaxed match respectively (F-score).
Third, the Venetian References corpus\footnote{\url{https://github.com/dhlab-epfl/LinkedBooksReferenceParsing}}~\cite{colavizza_annotated_2017} contains about 40,000 annotated bibliographic references from a corpus of books and journal articles on the history of Venice (19C-21C century) in Italian, English, French, German, Spanish and Latin.
Components of references (e.g. author, title, publication date, etc.) are annotated according to a custom tag set of 26 tags, and references themselves are classified according to the type of work they refer to (e.g. primary vs. secondary sources).

\subsubsection{Other}

We found one corpus in the domain of political writings. The De Gasperi corpus\footnote{\url{https://github.com/StefanoMenini/De-Gasperi-s-Corpus}}~\cite{tonelli_prendo_2019} consists of the complete collection of public documents by Alcide De Gasperi, Italy's Prime Minister in office from 1945 to 1953 and one of
the founding fathers of the European Union. This large corpus includes 2,762 documents published between 1901 and 1954 and belonging to a wide variety of genres. It was automatically annotated with parts of speech, lemmas, person and place names (by means of TextPro~\cite{pianta_textpro_2008}).
This corpus consists of clean texts extracted from the electronic versions of previously published volumes.
\subsection{Language representations}

As distributional representations, embeddings and language models need to be trained on large textual corpora in order to be effective. There exist several large-scale, diachronic collections of historical documents, such as the Europeana Newspaper collection~\cite{neudecker_making_2016}, the Trove Newspaper corpus~\cite{cassidy_publishing_2016}, the Digi corpus~\cite{kettunen_analyzing_2014}, and the \textit{impresso} public corpus~\cite{ehrmann_language_2020} (to mention but a few), which are now used to acquire historical language representations. Given their usefulness in many NLP tasks, embeddings and language models are increasingly shared by researchers, thus constituting a growing and quickly evolving pool of resources that can be used in historical NER. This section inventories existing historical language representations, an overview of which is given in Table \ref{tab:lexical-resources2}.

\subsubsection{Static embeddings}
As to traditional word embeddings, we could inventory two main resources. Sprugnoli et al.~\cite{sprugnoli_novel_2019} have released a collection of pre-trained word and sub-word English embeddings learned from a subset of the Corpus of Historical American English~\cite{davies_expanding_2012}, considering 37k texts published between 1860 and 1939 amounting to about 198 million words. These embeddings of 300 dimensions are available according to three types of word representations: embeddings based on linear bag-of-words contexts (GloVe~\cite{pennington_glove_2014}), on dependency parse-trees (Levy et al.~\cite{levy_dependencybased_2014}), and on bag of character n-grams (fastText \cite{bojanowski_enriching_2017}).\footnote{For the link to the published embeddings see \url{https://github.com/dhfbk/Histo}.} 
Doughman et al.~\citet{doughman_timeaware_2020} have created Arabic word embeddings from three Lebanese news archives, with materials published between 1933 and 2011.\footnote{Models as well as evaluation details can be found at: \url{https://doi.org/10.5281/zenodo.3538880}.} Archive-level as well as decade-level embeddings were trained using word2vec with a continuous bag of words model.
Given the imperfect OCRed, hyper-parameter tuning was used to maximise accuracy on a set of analogy tasks. 

Another set of traditional word embeddings consists of diachronic or dynamic embeddings, i.e. static embeddings trained on different time bins of a corpus and thereafter aligned according to different strategies (post-hoc alignment after training on different time bins, or incremental training). Such resources provide a view of words over time and are usually used in diachronic studies such as culturomics and semantic change, but can also be used to feed neural architectures for other tasks. Some of the pioneers in releasing such material were Hamilton et al.~\cite{hamilton_diachronic_2016}, who published a collection of diachronic word embeddings\footnote{\url{https://nlp.stanford.edu/projects/histwords/}} for English, French, German and Chinese, covering roughly 19C-20C. They were computed from many different corpora by using word2vec skip-gram with negative sampling. Later on, Hengchen et al.~\cite{hengchen_models_2019} released a set of diachronic embeddings of the same type in English, Dutch, Finnish and Swedish trained on large corpora of 19C-20C newspapers.\footnote{\url{https://zenodo.org/record/3270648}} More recently, Hengchen et al.~\cite{hengchen_collection_2021} pursued these efforts with the publication of diachronic word2vec and fastText models trained on a large corpus of Swedish OCRed newspapers (1645-1926) (the Kubhist 2 corpus, 5.5 billion tokens). Thanks to its ability to capture sub-word information, their fastText model allows for retrieving OCR misspellings and spelling variations, thus being a useful resource for post-OCR correction and historical normalisation.

\ra{1.3}
\begin{table*}[t]
\footnotesize
\centering
\resizebox{\linewidth}{!}{
\begin{tabular}{llllllllllll}
\textbf{Publication}  & \textbf{Type(s)} & \textbf{Model(s)}  & \textbf{Language(s)} & \textbf{Training Corpus} \\
\midrule
\citet{hamilton_diachronic_2016} & classic word embeddings & PPMI, SVD,  word2vec  & de, fr, en, cn & Google Books$+$COHA \\
Hengchen et al.~\cite{hengchen_models_2019} & classic word embeddings & word2vec  & en, nl, fi, se &  newspapers and periodicals \\
Hengchen et al.~\cite{hengchen_collection_2021} & char.-based word \& word embeddings & fastText, word2vec  & sv & Kubhist 2 & \\
Sprugnoli et al.~\cite{sprugnoli_novel_2019} & char.-based word \& word embeddings & dependency-based, fastText, GloVe & en & CHAE & \\
\citet{doughman_timeaware_2020} & classic word embeddings & word2vec & ar & Lebanese News Archives \\
\citet{ehrmann_overview_2020,ehrmann_language_2020} & char.-based word \& char.-level LM embeddings  & fastText, flair & de,fr,en &  \textit{impresso} corpus & \\
\citet{hosseini_neural_2021} & all types & word2vec, fastText, flair, BERT & en &  Microsoft British Library corpus & \\
Schweter et al.~\cite{schweter_robust_2019} & character-level LM embeddings & BERT, ELECTRA & de, fr &  Europeana Newspaper corpus & \\
Bamman et al.~\cite{bamman_latin_2020} & word-level LM embeddings & BERT & la &  various Latin corpora & \\
\bottomrule
\end{tabular}
}
\caption{Overview of available word embeddings and LMs trained on historical corpora.}
\label{tab:lexical-resources2}
\vspace{-4mm}
\end{table*}

\subsubsection{Contextualised embeddings} Historical character-level LM embeddings are currently available for German, French, and English.
For historical German, Schweter et al.~\cite{schweter_robust_2019} have trained contextualised string embeddings (flair) on articles from two titles from the Europeana newspaper collection, the \textit{Hamburger Anzeiger} (about 741 million tokens, 1888-1945) and the \textit{Wiener Zeitung} (some 801 million tokens, 1703-1875). Resulting embeddings are part of the Flair library.\footnote{With the ID \texttt{de-historic-ha-X} (HHA) and  \texttt{de-historic-wz-X} (WZ) respectively.} 
Next, in the context of the HIPE-2020 shared task, fastText word embeddings and flair contextualised string embeddings were made available as auxiliary resources for participants.\footnote{Available at \href{https://files.ifi.uzh.ch/cl/siclemat/impresso/clef-hipe-2020/}{files.ifi.uzh.ch/impresso/clef-hipe-2020/} and on Zenodo platform under DOI 10.5281/zenodo.3706808; Flair embeddings were also integrated into the Flair framework: \url{https://github.com/flairNLP/flair}. CC BY-NC 4.0 license applies.} They were trained on newspaper materials in French, German and English, and cover roughly 18C-21C (full details in~\cite{ehrmann_overview_2020} and~\cite{ehrmann_language_2020}).
Similarly, \citet{hosseini_neural_2021} published a collection of static (word2vec, fastText) and contextualised embeddings (flair) trained on the Microsoft British Library (MBL) corpus. MBL is a large-scale corpus composed of 47,685 OCRed books in English (1760-1900) which cover a wide range of subject areas including philosophy, history, poetry and literature, for a total of approximately 5.1 billion tokens. For each architecture, authors released models trained either on the whole corpus or on books published before 1850. 

Word-level LM embeddings trained on historical data are available for Latin, French, German and English. Latin BERT is a LM for Latin trained on 640 million tokens spanning 22 centuries.\footnote{\url{https://github.com/dbamman/latin-bert}} In order to reach a sufficiently large volume of training material, a wide variety of datasets was employed including the Perseus Digital Library, the Latin Wikipedia (Vicipaedia), and Latin texts of the Internet Archive. 
Extrinsic evaluation of the model was performed on POS tagging and word sense disambiguation, for which Latin BERT demonstrated state-of-the-art results.
For historical German and French, \citet{schweter_europeana_2020} published BERT and ELECTRA models trained on two subsets of the Europeana newspapers corpus, consisting of 8 and 11 billion tokens for German and French respectively. The German models were evaluated on two historical NE datasets, on which the ELECTRA models over-performed the BERT ones, leading to an overall improvement on the current state-of-the-art results reported by \citet{schweter_robust_2019}. Finally, for 19C English, BERT-based language models trained on the MBL corpus are available in the histLM model collection~\cite{hosseini_neural_2021}. One model was trained on the entire corpus, and additional models were created for different time slices to enable the study of linguistic and cultural changes over the 19C, by fine-tuning an existing contemporary model (BERT base uncased).

\vspace{0.15cm}
\noindent \textbf{Conclusion on Resources.} Resources for historical NER are not numerous but do exist. A few typologies and guidelines adapted for historical OCRed texts were published. More and more annotated corpora are being released, but the 17 that we could inventory here are far from the 121 inventoried in~\cite{ehrmann_named_2016} for modern NE processing. They are to a large extent built from historical newspaper collections, a type of document massively digitised during the last years. If historical newspaper contents lend themselves particularly well to NER, this preponderance could also be taken as an early warning of the risk of reproducing the news bias already observed for contemporary NLP~\cite{plank_what_2016}. Besides, NE-annotated historical corpora show a modest degree of multilingualism, and most of them are published under open licenses. As per language representations, historical embeddings and language models are not numerous but multiply rapidly.

\section{Approaches to Historical NER}
\label{sec:approaches}

This section provides an overview of existing work on NER for historical documents, organised by type of approach: rule-based, traditional machine learning and deep learning. The emphasis here is more on the implementation and settings of historical NER methods, while strategies to deal with specific challenges---regardless of the method---are presented in Section \ref{sec:strategies}. Since research was almost exclusively done in the context of individual projects, and since there was no shared gold standard up to recently, system performances are often not comparable. 
We therefore report results only when computed on sufficiently large data and explicitly state when results are comparable. All works deal with OCRed material unless mentioned otherwise. In absence of obvious thematic or technical grouping criteria, they are presented in order of publication (oldest to newest). Table \ref{tab:works} presents a synthetic view of the reviewed literature.

\subsection{Rule-based approaches}
\label{sec:rba}

As for modern NER, first NER works dealing with historical documents were mainly symbolic. Rule-based systems do not require training data and are easily interpretable, but need time and expertise for designing the rules. Numerous rule-based systems have been developed for modern NER, and they usually obtain good results on well-formed texts (see Section \ref{sec:RB}).

Early work performed NER over historical collections using the GATE language technology environment~\cite{cunningham_gate_2002}, which supports the manual creation of rules and gazetteers. Those work do not include formal evaluations but are worth mentioning as early exploration efforts, e.g. the adaptation of rules and gazetteers by Bontcheva et al.~\cite{bontcheva_using_2002} to recognise \textit{Person}, \textit{Location}, \textit{Occupation} and \textit{Status} entity types in 18C English court trials.
Among other difficulties, authors mention historical occupation names not present in gazetteers, orthographic variations (punctuation, spelling, capitalisation), and person name abbreviations.

Thereafter, most systems relied on custom rule sets and made substantial use of gazetteers, with the objective of addressing the domain and language peculiarities of historical documents. Jones et al.~\cite{jones_challenge_2006} designed a rule-based system to extract named entities from the Civil War years (1861-1865) of the American newspaper the \textit{Richmond Times Dispatch} (on manually segmented and transcribed issues). They focus on 10 entity types, some of them specific to the period and the material at hand such as warships, military units and regiments. Their system consists of three main phases: gazetteer lookup to extract easily identifiable entities; application of high precision rules to guess new names; and learning of frequency-based rules (e.g. how often \textit{Washington} appears as a person rather than a place, and in which context). Best results are obtained for \textit{Location} and \textit{Date}, while the identification of \textit{Person}, \textit{Organisation} and \textit{Newspaper titles} is lower. Based on a thorough error analysis, authors conclude that shorter but historically relevant gazetteers may be better than long ones, and make a plea for the development of comprehensive domain-specific knowledge resources.

Working on Swedish literary classics from the 19C, Borin et al.~\cite{borin_naming_2007} designed a system made of multiple modules: a gazetteer lookup and finite-state grammars module to recognise entities, a name similarity module to address lexical variation, and a document centred module to propagate labels based on documents' global evidence. They focused on 8 entity types and evaluated system modules' performances on an incremental basis. On all types together, the best F-measure reaches $89\%$, and recall is systematically lower than precision in all evaluation iterations (evaluation setting is partial match). The main sources of error are spelling variations, unknown names, and noisy word segmentation due to hyphenation in the original document.

Grover et al.~\cite{grover_named_2008} focused on two subsets of the Journal of the House of Lords of Great Britain, one from the late 17C and the other from early 19C, OCRed with different systems and at different times. OCR quality is erratic, and suffers from numerous quotation marks as well as from the presence of marginalia and of text portions in Latin. An in-house rule-based system, consisting of a set of rules applied incrementally with access to a variety of lexica, is applied to recognise person and place names. 
Before NE tagging, the system attempts to identify marginalia words and noisy characters in order to ignore them during parsing. 
The overall performance is evaluated against test sets of each period, which comprise significantly more person than location names. Results are comparable for person names for both 17C and 19C sets (ca. $75\%$ F-score), but the earliest period has significantly worse performance for locations ($24.1\%$ and $66.5\%$). In most configurations, precision is slightly above recall (evaluation setting not specified, most likely exact match). An error analysis revealed that character misspellings and segmentation errors (broken NEs) were the main factors impacting performances.

The experiments conducted by Broux et al.~\cite{broux_developing_2015} 
are part of an initiative aiming at improving access to texts from the ancient world. Working with a large collection of documentary texts produced between 800 BCE and 800 CE, including all languages and scripts written on any surface (mainly papyrological and epigraphical resources), one of the objective is to develop and curate onomastic lists and prosopographies of non-royal individuals attested as living during this period.\footnote{\textit{Onomastic} relates to the study of the history and origin of proper names (Oxford English dictionary), and \textit{prosopography} relates to the collection and study of information about a person.} Authors apply a rule-based system benefiting from a huge onomastic gazetteer covering names, name variants and morphological variants in several ancient languages and scripts. Rules encode various sets of onomastic patterns specific to Greek, Latin and Egyptian (Greek names are `simpler' than the often multiple Roman names, e.g. \textit{Gaius Iulius Caesar}) and specifically designed to capture genealogical information. This system is used to speed up manual NE annotation of texts, which in turn is used for network analysis in order to assist the creation of prosopographies. No formal evaluation is provided.

Fast-forwarding to contemporary times, Kettunen et al.~\cite{kettunen_names_2017} experimented with NER on a collection of Finnish historical newspapers from late 19C - early 20C. Authors insist on the overall poor quality of the OCR (word level correctness around $70\%-75\%$), as well as on the fact that they use an existing rule-based system designed for modern Finnish with no adaptation. Not surprisingly, this combination leads to rather low results with F-scores ranging from $30\%$ to $45\%$ for the 8 targeted entity types (evaluation setting is exact match). 
The main sources of errors are bad OCR and multi-word entities.

A recent work by Platas et al.~\cite{platas_medieval_2020} focuses on a set of manually transcribed Medieval Spanish texts (12C-15C) covering various genres such as legal documents, epic poetry, narrative, or drama. Based on the needs of literary scholars and historians, the authors defined a custom entity typology of 8 main types (plus sub-types). It covers traditional but also more specific types for the identification of name parts, especially relevant for Medieval Spanish person names featuring many attributes and complex syntactic structures (\textit{Don Alfonso por la gracia de Dios rey de Castiella de Toledo de Leon de Gallizia de Seuilla de Cordoua de Murcia e de Jaen}). The system is composed of several modules dedicated to recognising names using rules and/or gazetteers, increasing the coverage using variant generation and matching, and recognising person attributes using dependency parsing. Evaluated on a manually annotated corpus representative of the time periods and genres of the collection, 
the system reached satisfactory results with an overall F-score of $77\%$, ranging from $74\%$ to $87\%$ depending on the entity type (evaluation setting is exact match). As usual, recall is lower than precision, but differences are not high. Although these numbers are lower than what neural-based systems can achieve, this demonstrates the capacities and suitability of a carefully designed rule-based system.

Finally, it is also worth mentioning a series of work on the geoparsing of historical and literary texts. 
With the aim of analysing the interplay between geographical and fictional landscapes, \citet{moncla_automated_2017} experimented with a rule-based system relying on extensive gazetteers to recognise names of streets, houses, bridges, etc. in French Parisian novels from the 19C. With spatial entities featuring a high degree of regularity, the system reached very good results on a relatively small test set (evaluation settings are not entirely clear). Adapting the existing Edinburgh Geoparser system (derived from \citet{grover_named_2008} above) for historical texts, Alex et al.~\cite{alex_adapting_2015} carried out experiments to recognise place names in different types of 19C British historical documents. 
Besides the impact of OCR errors, main observations are that it is essential to perform place and person name recognition in tandem in order to better handle homonyms---even when dealing with place names only---, and that gazetteers need substantial adaptation, with careful switching on and off of standard vs domain-specific lexica. This system was also applied on a set of historical Edinburgh-specific documents, this time targeting fine-grained location names and considering three types of material: OCRed documents from 19C British novels,
manually crowd-corrected OCRed texts from the Project Gutenberg collection, and contemporary (born-digital) texts from Scottish authors~\cite{alex_geoparsing_2019}. Not surprisingly, place name recognition performs best on contemporary texts (but remains low with an F-score of $75\%$), worst on historical OCRed text (F-score $68\%$), and roughly in-between on crowd-corrected OCRed documents (F-score $72\%$). Precision scores are similar across the three collections, but recall scores vary considerably. Much research has been done on the geoparsing of cultural heritage material 
but is not further surveyed here.

\vspace{0.15cm}
\noindent \textbf{Conclusion on rule-based approaches}. Symbolic approaches were applied on a large variety of document types, domains and time periods (see Table~\ref{tab:works} for an overview of characteristics). In general, rule-based systems are modular and almost systematically include gazetteer lookup, rule incremental application, and variant matching. They have difficulties dealing with noisy and historical input, for which they require normalisation rules and additional linguistic knowledge. The number of work we could inventory, from the beginning of the 2000s until today, confirms the long-standing need for NER on historical documents as well as the suitability of symbolic approaches that can be better dealt with by non experts. Research nevertheless moved away from such systems in favour of machine learning ones.

\subsection{Traditional Machine Learning Approaches} 
\label{mla}

Machine learning algorithms inductively learn statistical models from annotated data on the basis of manually selected features (see Section~\ref{sec:ML}). Heavily researched and applied in the 2000s,  machine learning-based approaches contributed strong baselines for mainstream NER, and were rapidly adopted for NER on historical documents. In this section we review the usage of such traditional, pre-neural machine learning approaches on historical material, first considering works which apply already existing models, second which train new ones.

\subsubsection{Applying existing models} 

Early achievements adopted the `off-the-shelf' strategy with the application of pre-trained NER systems or web services to various historical documents, mainly with the objectives of assessing baselines and/or comparing system performances. This is the case of Rodriquez et al.~\cite{rodriquez_comparison_2012}, who compared the performances of four NER systems (Stanford CRF classifier, OpenNLP, AlchemyAPI, and OpenCalais) on two English datasets related to WWII: individual Holocaust survivor testimonies from the Wiener Library of London and letters of soldiers from King’s College archive. Evaluated on a small dataset, the recognition of \textit{Person}, \textit{Location} and \textit{Organization} reached an F-score between $47\%$ and $54\%$ for the testimonies (Stanford CRF being the most accurate), and between $32\%$ and $36\%$ for the  letters (OpenCalais performing best). Surprisingly, running the same evaluation on manually corrected OCR did not improve results significantly. Major sources of errors were different ways of naming and metonymy phenomena (e.g. warships named after people), and lack of background knowledge, especially for organisations.

\ra{1.2}
\begin{table*}[!t] 
\scriptsize
\centering
\begin{tabular}{rllllll}
\textbf{Publication} & \textbf{Domain} & \textbf{Document type} & \textbf{Time period} & \textbf{Language(s)} & \textbf{System} & \textbf{Comp.}   \\
\midrule
\addlinespace[0.15cm]
\multicolumn{7}{l}{\textbf{Rule-based}}\\
\addlinespace[0.05cm]

Bontcheva et al.~\cite{bontcheva_using_2002}      & legal       & court trials         & 18C        & en-GB  &  rule-based  &   \\
Jones et al.~\cite{jones_challenge_2006}          & news        & newspapers           & mid 19C    & en-US  &  rule-based  &   \\
\citet{borin_naming_2007}                         & literature  & literary classics    & 19C        & sv       & rule-based       &   \\
\citet{grover_named_2008}                         & state       & parliamentary proc.  & 17C \& 19C & en-GB  &  rule-based     &   \\
\citet{broux_developing_2015}                     & state       & papyri               & 4C-1C~\textsc{bce}   & egy,~el,~la  & lookup       &   \\
Kettunen et al.~\cite{kettunen_names_2017}        & news        & newspapers           & 19C-20C    & fi       & rule-based       &   \\
\citet{alex_adapting_2015}                        & state/literature  & parl. proc./classics   &  var & en-scotland    & lookup  &   \\
\citet{alex_geoparsing_2019}                      & literature  & novels               & 19C          & en-scotland     & lookup  &   \\
\citet{moncla_automated_2017}                     & literature  & novels               & 19C        & fr      & lookup  &   \\
Platas et al.~\cite{platas_medieval_2020}         & literature  & poetry, drama        & 12C-15C    & es     & rule-based       &   \\

\addlinespace[0.25cm]
\multicolumn{7}{l}{\textbf{Traditional machine learning}}\\
\addlinespace[0.05cm]

\citet{nissim_recognising_2004}                   & admin      & parish registers     & 18C-19C    & en-scotland &  MaxEnt    &     \\
\citet{packer_extracting_2010}                    & mix        & various               &  -      &   en         &  ensemble &   \\
\citet{rodriquez_comparison_2012}                 & egodocs    & letters \& testimonies & WWII     & en-GB      &  several    &   \\
\citet{galibert_extended_2012}                    & news       & newspapers           & 19C        & fr         &  several       &  \\ 
Dinarelli et al.~\cite{dinarelli_treestructured_2012} & news   & newspapers           & 19C        & fr           & CRF+PCFG  &   \\
Ritze et al.~\cite{ritze_named_2014}             & state       & admiralty court rec. & 17C        & en-GB      &  CRF &   \\
\citet{neudecker_largescale_2014}                & news        & newspapers           & 19C-20C    & de,~fr,~nl  &  CRF &   \\
Passaro et al.~\cite{passaro_piave_2014}         & state       & war bulletins        & 20C        & it           &  CRF &   \\
Kim et al.~\cite{kim_finding_2015}               & news        & newspapers           & -          & en           &  CRF       &   \\
\citet{ehrmann_diachronic_2016}                  & news        & newspapers           & 19C-20C    & fr           &  several  & \\ 
\citet{aguilar_named_2016}                       & news        & medieval charters    & 10C-13C    & la  &  CRF      &   \\
\citet{erdmann_challenges_2016}                  & literature           & classical texts &                    1C~\textsc{bce}-2C          & la          &  CRF      &   \\
Ruokolainen et al.~\cite{ruokolainen_recherche_2018}  & news   & newspapers           & 19C-20C    & fi          &  CRF+gaz   &   \\
\citet{won_ensemble_2018}                        &  egodocs    & letters              &  17-18C      &  en-GB    &  ensemble &   \\
\citet{elvaigh_irisa_2020}                       & news        & newspapers~(\textsc{hipe})& 19C-20C & de,~en,~fr   &  CRF  \\
Kogkitsidou et al.~\cite{kogkitsidou_normalisation_2020}  & literature &  theatre and memoirs  &   16C-17C         & French   &  several     &   \\

\addlinespace[0.25cm]
\multicolumn{7}{l}{\textbf{Deep Learning}}\\
\addlinespace[0.05cm]

Riedl et al.~\cite{riedl_named_2018}                & news         & newspapers         & 19C-20C     &  de        & BiLSTM-CRF & $\lozenge$  \\
Rodrigues A. et al.~\cite{rodriguesalves_deep_2018} & bibliometry  & journals \& monographs  & 19C-20C & multi     & BiLSTM-CRF &   \\
\citet{sprugnoli_arretium_2018}                     & literature   & travel writing     & 19C-20C     & en-US      & BiLSTM-CRF  &   \\
\citet{ahmed_biofid_2019}                           & biodiversity & scholarly pub.     & 19C-20C     &  de        & BiLSTM-CRF  &   \\
\citet{kew_geotagging_2019}                         & literature   & alpine texts       & 19C-20C     &  multi        & BiLSTM-CRF &  \\
Schweter et al.~\cite{schweter_robust_2019}         & news         & newspapers         & 19C-20C     &  de        & BiLSTM-CRF  &   $\lozenge$ \\
Labusch et al.~\cite{labusch_bert_2019}             & news         & newspapers         & 19C-20C     &  de        & BERT & $\lozenge$  \\
\citet{dekhili_hybrid_2020}                         &  news        & newspapers~(\textsc{hipe})               & 19C-20C   & fr            & BiLSTM-CRF &  $\blacklozenge$ \\
Ortiz S. et al.~\cite{ortizsuarez_sinner_2020}      &  news        & newspapers~(\textsc{hipe})& 19C-20C   & fr, de   & BiLSTM-CRF      &  $\blacklozenge$ \\
Kristanti et al.~\cite{kristanti_delft_2020}        &  news        & newspapers~(\textsc{hipe})& 19C-20C   & en, fr   & BiLSTM-CRF  &  $\blacklozenge$ \\
\citet{provatorova_named_2020}                      &  news        & newspapers~(\textsc{hipe})& 19C-20C   & de,~en,~fr   & BiLSTM-CRF      &  $\blacklozenge$ \\
Todorov et al.~\cite{todorov_transfer_2020}          &  news        & newspapers~(\textsc{hipe})& 19C-20C   & de,~en,~fr   & BiLSTM-CRF  &  $\blacklozenge$ \\
Schweter et al.~\cite{schweter_triple_2020}         &  news        & newspapers~(\textsc{hipe})& 19C-20C   & de   & BiLSTM-CRF      &  $\blacklozenge$\\
Labusch et al.~\cite{labusch_named_2020}            &  news        & newspapers~(\textsc{hipe})& 19C-20C   & de,~en,~fr   & BERT  &  $\blacklozenge$ \\
\citet{ghannay_experiments_2020}                    &  news        & newspapers~(\textsc{hipe})& 19C-20C   & fr   &       &  $\blacklozenge$ \\
\citet{boros_robust_2020}                           &  news        & newspapers~(\textsc{hipe})& 19C-20C   & de,~en,~fr   & BERT  &  $\blacklozenge$ \\
\citet{swaileh_named_2020}                          &  economy     & financial yearbooks       & 20C       & de, fr  &  BiLSTM-CRF &   \\
Yu et al.~\cite{yu_bertbased_2020}                  &  history     & state official books      & 1\textsc{bce}-17C  & zh  &  BERT &   \\
\citet{hubkova_czech_2020}                          &  news        & newspapers                & 19C-20C   & cz            & BiLSTM &   \\
\bottomrule

\end{tabular}
\caption{Historical NER literature overview. Papers are grouped by family of approaches and ordered by publication year. `\textsc{Comp.}' stands for comparable and denotes works whose results are obtained on same test sets.}
\label{tab:works}
\end{table*}

Along the same line, Ehrmann et al.~\cite{ehrmann_diachronic_2016} conducted experiments on French historical newspapers on a diachronic basis (covering 200 years) for the types \textit{Person} and \textit{Location}, with the objective of investigating whether NER performance degrades when going back in time. Their study includes four systems representative of major approaches for NER: a rule-based system, a supervised machine learning one (MaxEnt classifier), and two proprietary web services offering NER functionalities (AlchemyAPI and DandelionAPI). They showed that, compared to a baseline on contemporary news, all systems feature degraded performances, both in absolute terms and over time (maximum of $67.6\%$ F-score for person names for the best system, with exact match). As for time-based observation, precision is quite irregular, with several ups and downs for all systems for both entity types, but recall shows less variability and a slight but regular increase for \textit{Person}, suggesting that person names are less stable than location names and therefore better recognised when more recent. 

Focusing on the impact of historical language normalisation (in this respect see also Section \ref{sec:strategy-dia}), Kogkitsidou et al.~\cite{kogkitsidou_normalisation_2020} also used and benchmarked several systems (rule-based and machine learning) for the recognition of \textit{Location} names in French literary texts from the 16C and 17C. When applied without any adaptation, systems features very diverse performances, from very low ($36\%$) to reasonable ($70\%$) F-scores, with rule-based ones being better at precision, and machine learning ones at recall.

Ritze et al.~\cite{ritze_named_2014} worked on historical records of the English High Court of Admiralty of the 17C and used the Stanford CRF classifier with its default English model to recognise \textit{Person} and \textit{Location} types (others were considered but not evaluated). Given the very specific domain of this corpus, obtained results were reasonable, with a precision in the $77\%$ for both types (recall was not reported).

Finally, some adopt the approach of ensembling systems, i.e. of considering NE predictions not from one but several recognisers, according to various voting strategies. Packer et al.~\cite{packer_extracting_2010} applied three algorithms (dictionary-based, regular expressions-based, and HMM-based) in isolation and in combination for the recognition of person names in various types of English OCRed documents. They observed increased performances (particularly a better P/R balance) with a majority vote ensembling. 
Won et al.~\cite{won_ensemble_2018} worked on British personal archives 
from 16C and 17C and applied five different systems to recognise place names. They too observed that the combination of multiple systems through a majority vote (with a minimum of two to a maximum of three votes) was able to consistently outperform the individual NER systems.

Mere application of existing systems, these work illustrate the inadequacy of already trained NER models for historical texts. 
Performances (and settings) of these baseline studies are extremely diverse, but the following constants are observed: recall is always the most affected, and the \textit{Location} type is usually the most robust.

\subsubsection{Training models} 

Other work trained NER systems anew on custom material. Early attempts include the experiments of Nissim et al.~\cite{nissim_recognising_2004} on \textit{Location} entity type in manually transcribed Scottish parish registers of the late 18C and early 19C. They trained a maximum entropy tagger with its in-built standard features 
on a dataset of ca. 6000 location mentions and obtained very satisfying performance ($94.2\%$ F-score), which they explained by the custom training data and the binary classification task (location vs non-location).

Subsequently, the most frequently used system is the Stanford CRF classifier\footnote{https://nlp.stanford.edu/software/CRF-NER.html}\cite{finkel_incorporating_2005}, particularly on historical newspapers. Working with the press collection of the National Library of Australia, Kim et al.~\cite{kim_finding_2015} evaluated two Stanford CRF models, the default English one trained on CoNLL-03 English data, and a custom one trained on 600 articles of the Trove collection (the time period of the sample is not specified). Interestingly, the model trained on in-domain data did not outperform the default one, and both yielded F-scores around $75\%$ for \textit{Person} and \textit{Location}, with a drop below $50\%$ for \textit{Organisation}. Neudecker et al.~\cite{neudecker_largescale_2014} focused on newspaper material in French, German and Dutch from the Europeana collection~\cite{neudecker_open_2016}, on which they trained a Stanford CRF model with additional gazetteers. The 4-fold cross-evaluation yielded F-scores in the range of 70-80\% for Dutch and French, while no results were reported for German. For both languages, recall was significantly lower than precision. Working on Finnish historical newspapers, Ruokolainen et al.~\cite{ruokolainen_recherche_2018} considered \textit{Person} and \textit{Location} and trained the Stanford CRF classifier on manually corrected OCRed material, with large gazetteers covering inflected forms. The model gave satisfying performances with F-scores of $87\%$ (location) and $80\%$ (person) on a test set taken from the same manually corrected data, and of $78\%$ and $71\%$ on non-corrected OCR texts (with recall being lower than precision). This time on French, and taking advantage of the Quaero Old Press corpus, Galibert et al.~\cite{galibert_extended_2012} organised a small evaluation campaign where three anonymous systems participated. 
Stochastic systems performed best (especially on noisy entities), with an F-score of $65.2\%$ across all types (person, location and organisation). Also working on French newspapers in the context of HIPE-2020, Elvaigh et al.~\cite{elvaigh_irisa_2020} (slightly) fine-tuned the CRF baseline provided by the organisers and reached $66\%$ on all types (exact match), two points more than the baseline.

Going back in time, Aguilar et al.~\cite{aguilar_named_2016} experimented NER on manually transcribed 
Latin medieval charters 
from the 10C to 13C. 
Focusing on person and place names, they used dedicated pre-processing and trained a CRF classifier using the Wapiti toolkit.\footnote{\url{https://wapiti.limsi.fr/}} Results are remarkable, on average in the $90\%$ for both types, certainly due to the regularity of the documents in terms of names, naming conventions, context and overall structure.

Finally, Passaro et al.~\cite{passaro_piave_2014} attempted to extract entities from WWI and WWII Italian official war bulletins. 
They focused on the traditional entity types, plus \textit{Military organisations}, \textit{Ships} and \textit{Airplanes}. The Stanford system was trained (without gazetteers) on semi-automatically annotated data from the two periods as well as on contemporary Italian news, and various experiments mixing in- vs. out-of-time data were carried out. Results showed that performances are highest when the model is trained on data close in time, that entities of type \textit{Location} are systematically better recognised, and that custom types (ships, military organisations, etc.) are poorly recognised.

\vspace{0.15cm}
\noindent \textbf{Conclusion on traditional machine learning approaches.} Overall, the availability of machine learning-based NER systems that could either be applied as such or trained on new material greatly fostered a second wave of experiments on historical documents. Settings are quite diverse, and so are the performances, but F-scores are usually in the order of $60-70\%$, which is significantly lower than those usually obtained on contemporary material (frequently in the $90\%$). The Stanford CRF classifier is by far the most commonly used, as well as CRF in general. Not surprisingly, performances are higher when systems are trained on in-domain material.

\subsection{Deep Learning Approaches}
\label{dla}

Latest developments in historical NER are dominated by deep learning techniques which have recently shown state-of-the-art results for modern NER. Deep learning-based sequence labelling approaches rely on word and character distributed representations and learn sentence or sequence features during end-to-end training. Most models are based on BiLSTM architectures or self-attention networks, and use a CRF layer as tag decoder to capture dependencies between labels (see Section \ref{sec:DL}). Building on these results, much work attempt to apply and/or adapt deep learning approaches to historical documents, under different settings and following different strategies.

\subsubsection{Preliminary comments}

Let us begin with some observations on the main lines of research. In a feature learning context the crucial point is, by definition, the capacity of the model to learn or reuse appropriate knowledge for the task at hand. Given a situation of time and domain shifts and of resource scarcity, what is at stake for neural-based historical NER approaches is to capture historical language idiosyncrasies (including OCR noise) and to adequately leverage previously learned knowledge --- a process made increasingly possible with the usage of pre-trained language models in a transfer learning context. Transfer learning (TL) refers to a set of methods which aims at leveraging knowledge from a source setting and adapting it to a target setting~\cite{pan_survey_2010}. TL is not new in NLP but was recently given considerable momentum, in particular sequential transfer learning where the source task (e.g. language modeling) differs from the target task (e.g. NER). In this supervised TL setting, a widely used process is to first learn representations on a large unlabelled corpus (source), before adapting them to a specific task using labelled data (target). The previously learned model can be adapted to the target task in different ways, the most frequent being weight adaptation, where pre-trained weights are either kept unchanged (`frozen') and used as features in the downstream model (feature extraction), or fine-tuned to the target task and used as initialisation in the downstream model (fine-tuning)~\cite{ruder_transfer_2019}.

To date, most DL approaches to historical NER have primarily focused on experimenting with a) different input representations, that is to say embeddings of different granularity (character, sub-word, word), learned at the type or token level (static vs. contextualised) and derived from domain data or not (in vs. out-of-domain), and b) different transfer learning strategies. Those aspects are often intermingled in various experiments reported in the literature, which does not easily lend itself to a clear-cut narrative outline. The discussion which follows is organised according to the demarcation line `words vs. words-in-context', complemented with observations on TL settings and types of networks. However imperfect this line is, it reflects the recent evolution of incorporating more context and of testing all-round language models in historical settings. As a complement, and in order to frame further the discussion, we identified a set of key research questions from the types of experiments reported in publications, summarised in Table \ref{tab:DL-exp}. 

\ra{1.2}
\begin{table*}[t] 
\footnotesize
\centering
\begin{tabular}{@{}lll@{}}
\textbf{Research questions} & \textbf{Experiments} & \textbf{Publication} \\
\toprule
\addlinespace[0.1cm]
\textbf{Input representation} & &\\
\multicolumn{2}{l}{\textit{Which type of embedding is best?}} & \\
\addlinespace[0.15cm]
& Test different static embedding algorithms & \cite{sprugnoli_arretium_2018} \\
& Test different static embedding granularity  & \cite{rodriguesalves_deep_2018} \\
& Use modern static embeddings (word2vec, fastText) & \cite{hubkova_czech_2020} \\
& Use modern char-level LM embeddings (Flair) & \cite{swaileh_named_2020} \\
& Use modern word-level LM embeddings (BERT, ELMo) &  \cite{yu_bertbased_2020,provatorova_named_2020,ghannay_experiments_2020}\\
& Uses stack of modern embeddings  &  \cite{rodriguesalves_deep_2018,kristanti_delft_2020,ortizsuarez_sinner_2020}\\

\addlinespace[0.22cm]
\textbf{Transfer learning} & & \\
\multicolumn{2}{l}{\textit{How well modern embeddings can transfer to historical texts?}} & \\
\multicolumn{2}{l}{\textit{What is the impact of in-domain embeddings?}} & \\
\multicolumn{2}{l}{\textit{Is more task-specific labelled data more helpful than big or in-domain LMs?}} & \\
\addlinespace[0.15cm]
& Test modern vs. historical static embeddings & \cite{riedl_named_2018} \\
& Test modern vs. historical char-level LM embeddings & \cite{kew_geotagging_2019,schweter_robust_2019,dekhili_hybrid_2020,ortizsuarez_sinner_2020,schweter_triple_2020} \\
& Test modern vs. historical word-level LM embeddings &  \cite{ahmed_biofid_2019,schweter_robust_2019,labusch_bert_2019}\\
& Test stack of embeddings & \cite{schweter_robust_2019,ahmed_biofid_2019,schweter_triple_2020,labusch_bert_2019,todorov_transfer_2020,boros_robust_2020} \\
& Test feature extraction (frozen) vs. fine-tuning & \cite{rodriguesalves_deep_2018,hubkova_czech_2020,provatorova_named_2020} \\
& Test different training corpus sizes & \cite{riedl_named_2018,ahmed_biofid_2019,kristanti_delft_2020} \\
& Test cross-corpus model application  & \cite{riedl_named_2018,labusch_bert_2019,todorov_transfer_2020,kristanti_delft_2020,boros_robust_2020} \\
& Test cross-corpus model training  & \cite{riedl_named_2018} \\

\addlinespace[0.22cm]
\textbf{Neural architecture} & & \\
\multicolumn{2}{l}{\textit{How neural approaches compare to traditional CRFs? }} & \\
\multicolumn{2}{l}{\textit{What is the best neural architecture with which decoder?}} & \\
\addlinespace[0.15cm]
& Compare BiLSTM and traditional CRF &  \cite{sprugnoli_arretium_2018,riedl_named_2018,rodriguesalves_deep_2018,ortizsuarez_sinner_2020}\\
& Compare CRF decoder vs. softmax decoder & \cite{rodriguesalves_deep_2018} \\
& Compare BiLSTM and LSTM & \cite{hubkova_czech_2020} \\
& Test single vs. multitask learning & \cite{rodriguesalves_deep_2018,todorov_transfer_2020}\\
& Compare transformers and BiLSTM & \cite{boros_robust_2020} \\
\bottomrule
\end{tabular}
\caption{Synthetic view of DL experiments mapped with research questions.}
\label{tab:DL-exp}
\vspace{-4mm}
\end{table*}

\subsubsection{Approaches based on static embeddings}

First attempts 
are based on state-of-the-art BiLSTM-CRF 
and investigate the transferability of various types of pre-trained static embeddings to historical material. They all use traditional CRFs as baseline.

Focusing on location names in 19-20C English travelogues,\footnote{Corpus presented in Section \ref{sec:lit}.} Sprugnoli~\cite{sprugnoli_arretium_2018} compares two classifiers, Stanford CRF and BiLSTM-CRF, and experiment with different word embeddings: GloVe embeddings, based on linear bag-of-words contexts and trained on Common Crawl data ~\cite{pennington_glove_2014}, Levy  and  Goldberg  embeddings, produced  from the  English  Wikipedia  with  a dependency-based approach~\cite{levy_dependencybased_2014}, and fastText embeddings, also trained on the  English  Wikipedia but using sub-word information~\cite{bojanowski_enriching_2017}. Additionally to these pre-trained vectors, Sprugnoli trains each embedding type afresh on historical data (a subset of the Corpus of Historical American English), ending up with 3$\times$2 input options for the neural model. Both classifiers are trained on a relatively small labelled corpus. Results show that the neural approach performs systematically and remarkably better than CRF, with a difference ranging from 11 to 14 F-score percentage points, depending on the word vectors used (best F-score is 87.4$\%$). If in-domain supervised training improves the F-score of the Stanford CRF module, it is worth noting that the gain is mainly due to recall, the precision of the English default model remaining higher. In this regard, the neural approach shows a better P/R balance across all settings. With respect to embeddings, linear bag-of-words contexts (GloVe) prove to be more appropriate (at least in this context), with its historical embeddings yielding the highest scores across all metrics (fastText following immediately after). A detailed examination of results reveals an uneven impact of in-domain embeddings, leading either to higher precision but lower recall (Levy and GloVe), or higher recall but lower precision (fastText and GloVe). Overall, this work shows the positive impact of in-domain training data: the BiLSTM-CRF approach, combined with in-domain training set and in-domain historical embeddings, systematically outperforms the linear CRF classifier.

In the context of reference mining in the arts and humanities, Rodriguez et al.~\cite{rodriguesalves_deep_2018} also investigate the benefit of BiLSTM over traditional CRFs, and of multiple input representations. Their experiments focus on three architectural components: input layer (word and character-level word embeddings), prediction layers (Softmax and CRF), and learning setting (multi-task and single-task). Authors consider a domain-specific tagset of 27 entity types covering reference components (e.g. author, title, archive, publisher) and work with 19-21C scholarly books and journals featuring a wide variety of referencing styles and sources.\footnote{Corpus presented in Section~\ref{sec:lit}} While character-level word embeddings, likely to help with OCR noise and rare words, are learned either via CNN or BiLSTM, word embeddings are based on word2vec and are tested under various settings: present or not, pre-trained on the in-domain raw corpus or randomly initialised, and frozen or fined-tuned on the labelled corpus during training. Among those settings, the one including in-domain word embeddings further fine-tuned during training and CRF prediction layer yields the best results ($89.7\%$ F-score). Character-level embeddings provide a minor yet positive contribution, and are better learned via BiLSTM than with CNN. The BiLSTM architecture outperforms the CRF baseline by a large margin (+$7\%$), except for very infrequent tags. Overall, this work confirms the importance of word information (rather in-domain, though here results with generic embeddings were not reported) and the remarkable capacities of a BiLSTM network to learn features, better decoded by a CRF classifier than a softmax function.

Working with Czech historical newspapers,\footnote{Corpus presented in Section~\ref{sec:news}} Hubková et al.~\cite{hubkova_czech_2020} target the recognition of five generic entity types. Authors experiment with two neural architectures, LSTM and BiLSTM, followed by a softmax layer. Both are trained on a relatively small labelled corpus (4k entities) and fed with modern fastText embeddings (as released by the fastText library) under three scenarios: randomly initialised, frozen, and fine-tuned. Character-level word embeddings are not used. Results show that the BiLSTM model based on pre-trained embeddings with no further fine-tuning performs best ($73\%$ F-score). Authors do not comment on the performance degradation resulting from fine-tuning, but one reason might be the small size of the training data.

Rather than aiming at calibrating a system to a specific historical setting, Riedl et al.~\cite{riedl_named_2018} adopt a more generic stance and investigate the possibility of building a German NER system that performs at the state of the art for both contemporary and historical texts. The underlying question---whether one type of model can be optimised to perform well across settings--- naturally resonates with the needs of cultural heritage institution practitioners (see also Schweter et al.~\cite{schweter_robust_2019} and Labush et al.~\cite{labusch_bert_2019} hereafter). Experimental settings consist of: two sets of German labelled corpora, with large contemporary datasets (CoNNL-03 and GermEval) and small historical ones (from the Friedrich Temann and Austrian National library); two types of classifiers, CRFs (Stanford and GermaNER) and BiLSTM-CRF; finally, for the neural system, usage of fastText embeddings derived from generic (Wikipedia) and in-domain (Europeana corpus) data. On this base, authors perform three experiments. The first investigates the performances of the two types of systems on the contemporary datasets. On both GermEval and CoNNL, the BiLSTM-CRF models outperform the traditional CRF ones, with Wikipedia-based embeddings yielding better results than the Europeana-based ones. It is noteworthy that the GermaNER CRF model performs better than the LSTM of Lample et al.~\cite{lample_neural_2016} on CoNLL-03, but suffers from low recall compared to BiLSTM. The second experiment focuses on all-corpora crossing, with each system being trained and evaluated on all possible combinations of contemporary and historical corpora pairs. With no surprise, best results are obtained when models are trained and evaluated on the same material. Interestingly, CRFs perform better than BiLSTM in the historical setting (i.e. train and test sets from historical corpora) by quite a margin, suggesting that although not optimised for historical texts, CRFs are more robust than BiLSTM when faced with small training datasets. The type of embeddings (Wikipedia vs. Europeana) plays a minor role in the BiLSTM performance in the historical setting. Ultimately, the third experiment explores how to overcome this neural net dependence on large data with domain adaptation transfer learning: the model is trained on a contemporary corpus until convergence and then further trained on a historical one for a few more epochs. Results show consistent benefits for BiLSTM on historical datasets (ca. +4 F-score percentage points). In general, main difficulties relate to OCR mistakes and wrongly hyphenated words due to line breaks, and to the \textit{Organisation} type. Overall, this work shows that BiLSTM and CRF achieve similar performances in a small-data historical setting, but that BiLSTM-CRF outperforms CRF when supplied with enough data or in a transfer learning setting.

\vspace{0.15cm}
This first set of work confirms the suitability of the state-of-the-art BiLSTM-CRF approach for historical documents, with the major advantage of not requiring feature engineering. Provided that there is enough in-domain training data, this architecture obtains better performances than traditional CRFs (the latter performing on par or better otherwise). In-domain pre-training of static word embeddings seems to contribute positively, although to various degrees depending on the experimental settings and embedding types. Sub-word information (either character embeddings or character-based word embeddings) also appears to have positive effect.

\subsubsection{Approaches based on character-level LM embeddings}

Approaches described above rely on static, token-level word representations which fail to capture context information. This drawback can be overcome by context-dependent representations derived from the task of modelling language, either as distribution over characters, such as the Flair contextual string embeddings~\cite{akbik_flair_2019}, or over words, such as BERT~\cite{devlin_bert_2019} and ELMo~\cite{peters_deep_2018} (see Section \ref{sec:embedd}). Such representations have boosted performances of modern NER and are also used in the context of historical texts. This section considers work based on character-based contextualised embeddings (flair).

In the context of the CLEF-HIPE-2020 shared task~\cite{ehrmann_extended_2020}, Dekhili et al.~\cite{dekhili_hybrid_2020} proposed different variations of a BiLSTM-CRF network, with and without the in-domain HIPE flair embeddings and/or an attention layer. The gains of adding one or the other or both are not easy to interpret, with uneven performances of the model variants across NE types. Their overall F-scores range from $62\%$ to $65\%$ under the strict evaluation regime. For some entity types the CRF baseline is better than the neural models, and the benefit of in-domain embeddings is overall more evident than the one of the attention layer (which proved more useful in handling metonymic entities).

\citet{kew_geotagging_2019} address the recognition of toponyms in an alpine heritage corpus consisting of over 150 years of mountaineering articles in five languages (mainly from the Swiss and British Alpine Clubs). Focusing on fine-grained entity types (city, mountain, glacier, valley, lake, and cabin), the authors compare three approaches. The first is a traditional gazetteer-based approach completed with a few heuristics which achieves high precision  across types ($88\%$ P, $73\%$ F-score), and even very high precision ($>95\%$) for infrequent categories with regular patterns. Suitable for reliable location-based search but suffering from low recall, this approach is then compared with a BiLSTM-CRF architecture. The neural system is fed with stacked embeddings composed of in-domain contextual string embeddings pre-trained on the alpine corpus concatenated with general-purpose fastText word embeddings pre-trained on web data, and trained on a silver training set containing 28k annotations obtained via the application of the gazetteer-based approach. The model leads to an increase of recall for the most frequent categories, without degrading precision scores ($76\%$ F-score). This shows the generalisation capacity of the neural approach in combination with context-sensitive string embeddings and given sufficient training data. Finally, authors experiment with crowd-corrected annotations and observe that already a small number of corrections on the silver data has a positive impact (+3 F-score percentage point).

\citet{swaileh_named_2020} target even more specific entity types in French and German financial yearbooks from the first half of 20C. They apply a BiLSTM-CRF network trained on custom data and fed with modern flair embeddings. Results are very good (between $85\%$ to $95\%$ F-score depending on the book sections), with the CRF baseline and the BiLSTM model performing on par for French books, and BiLSTM being better than CRF for the German one, which has a lower OCR quality. Overall, these performances can be explained by the regularity of the structure and language as well as the quality of the considered material, resulting in stable contexts and non-noisy entities.

\subsubsection{Approaches based on word-level LM embeddings}

The release of pre-trained contextualised language model-based word embeddings such as BERT (based on transformers) and ELMo (based on LSTM) pushed further the upper bound of modern NER performances. They show promising results either in replacement or in combination with other embedding types, and offer the possibility of being further fine-tuned~\cite{li_survey_2020}. If they are becoming a new paradigm of modern NER, the same seems to be true for historical NER.

\paragraph{Using pre-trained modern embeddings}
We first consider work based on pre-trained modern LM-based word embeddings (BERT or ELMo) without extensive comparison experiments. They make use of BiLSTM or transformer architectures. 

Working on the ``Chinese Twenty-Four Histories'', a set of Chinese official history books covering a period from 3000 BCE to 17C, Yu et al.~\cite{yu_bertbased_2020} face the problems of the complexity of classical Chinese and of the absence of appropriate training data in their attempt to recognise \textit{Person} and \textit{Location}. Their BiLSTM-CRF model is trained on a NE-annotated modern Chinese corpus and makes use of modern Chinese BERT embeddings in a feature extraction setting (frozen). Evaluated on a (small) dataset representative of the time span of the target corpus, the model achieves relatively good performances (from $72\%$ to $82\%$ F-score depending on the book), with a pretty good P/R balance, better results for \textit{Location} than for \textit{Person}, and on recent books. Given the complete `modern' setting of embeddings and training labelled data, those results shows the benefit of large LM-based embeddings---keeping in mind the small size of the test set and perhaps the regularity of entity occurrences in the material, not detailed in the paper.

Also based on the bare usage of state-of-the-art LM-based representations is a set of work from the HIPE-2020 evaluation campaign. These work tackle the recognition of five entity types in about 200 years of historical newspapers in French, English, and German.\footnote{Corpus presented in Section \ref{sec:news}.} The task included various NER settings, however only the coarse literal NE recognition is considered here. \citet{ortizsuarez_sinner_2020} focused on French and German. They first pre-process the newspaper line-based format (or column segments) into sentence-split segments before training a BiLSTM-CRF model using a combination of modern static fastText and contextualised ELMo embeddings as input representations. They favoured ELMo over BERT because of its capacity to handle long sequences and its dynamic vocabulary thanks to its CNN character embedding layer. In-domain fastText embeddings provided by the organisers were tested but performed lower. Their models ranked third on both languages during the shared task, with strict F-score of $79\%$ and $65\%$ for French and German respectively. The considerably lower performance of their improved CRF baseline illustrates the advantage of contextual embeddings-based neural models. Ablation experiments on sentence splitting showed an improvement of 3.5 F-score percentage points on French data (except for \textit{Location}) confirming the importance of proper context for NER neural tagging.

Running for French and English, Kristanti et al.~\cite{kristanti_delft_2020} also make use of a BiLSTM-CRF relying on modern fastText and ELMo emddings. In the absence of training set for English, authors use the CoNLL-2012 corpus, while for French the training data is further augmented with another NE-annotated journalistic corpus from 1990, which proved to have positive impact. They scored at $63\%$ and $52\%$ in terms of strict F-score for French and English respectively. Compared to the French results of Ortiz Su\`{a}ez et al., Kristanti et al. use the same French embeddings but a different implementation framework and different hyper-parameters, and does not apply sentence segmentation.

Finally, still within the HIPE-2020 context, two teams tested pre-trained LM embeddings with transformer-based architectures. \citet{provatorova_named_2020} proposed an approach based on the fine-tuning of BERT models using Huggingface's transformer framework for the three shared task's languages, using the cased multilingual BERT base model for French and German and the cased monolingual BERT base model for English. They used the CoNLL-03 data for training their English model, the HIPE data for the others, and additionally set up a majority vote ensemble of 5 fine-tuned model instances per language in order to improve the robustness of the approach. Their models achieved F-scores of $68\%$, $52\%$ and $47\%$ for French, German and English respectively. \citet{ghannay_experiments_2020} used CamemBERT, a multi-layer bidirectional transformer similar to ROBERTa~\cite{martin_camembert_2020,liu_roberta_2019} initialised with a pre-trained modern French CamemBERT and completed with a CRF tag decoder. This model obtained the second-best results for French with $81\%$ strict F-score.

\vspace{0.1cm}
Even when learned from modern data, pre-trained LM-based word embeddings encode rich prior knowledge that effectively support neural models trained on (usually) small historical training sets. As for HIPE-related systems, it should be noted that word-level LM embeddings systematically lead to slightly higher recall than precision, demonstrating their powerful generalisation capacities, even on noisy texts.

\paragraph{Using modern and historical pre-trained embeddings} As for static embeddings, it is logical to expect higher performances from LM-embeddings when pre-trained on historical data, in combination with modern ones or not. The set of work reviewed here explores this perspective.

\citet{ahmed_biofid_2019} work on the recognition of universal and domain-specific entities in German historical biodiversity literature.\footnote{Corpus presented in Section \ref{sec:lit}} They experiment with two BiLSTM-CRF implementations (their own and Flair framework) which both use modern token-level German word embeddings and are trained on the BIOfid corpus. Experiments consist in adding richer representations (modern Flair embeddings, additionally completed by newly trained ELMo embeddings or BERT base multilingual cased embeddings) or adding more task-specific training data (GermEval, CoNLL-03 and BIOfid). Models perform more or less equally, and authors explained the low gain of in-domain ELMo embdedings by the small size of the training data (100k sentences). Higher gains come with larger labelled data, however the absence of ablation tests hinders the complete understanding of the contribution of the historical part of this labelled data, and the use of two implementation frameworks does not warrant full results comparability.

Both Schweter et al.~\cite{schweter_robust_2019} and Labusch et al.~\cite{labusch_bert_2019} build on the work of Riedl et al.~\cite{riedl_named_2018} and try to improve NER performances on the same historical German evaluation datasets, thereby constituting (with HIPE-2020) one of the few sets of comparable experiments. Schweter et al. seek to offset the lack of training data by using only unlabelled data via pre-trained embeddings and language models. They use the Flair framework to train and combine (``stack'') their language models, and to train a BiLSTM-CRF model. Their first experiment consists in testing various static word representations, with: character embeddings learned during training, fastText embeddings pre-trained on Wikipedia or Common Crawl (with no sub-word information), and the combination of all of these. While Riedl et al. experimented with similar settings (character embeddings and pre-trained modern and historical fastText embeddings), it appears that combining Wikipedia and Common Crawl embeddings leads to better performances, even higher than the transfer learning setting of Riedl et al. using more labelled data. As a second experiment, Schweter et al. use pre-trained LM embeddings: flair embeddings newly trained on two historical corpora having temporal overlaps with the test data, and two modern pre-trained BERT models (multilingual and German). On both historical test sets, in-domain LMs yield the best results (outperforming those of Riedl et al.), all the more so when the temporal overlap between embedding and task-specific training data is large. This demonstrates that the selection of the language model corpus plays an important role, and that unlabelled data close in time might have more impact than more (and difficult to obtain) labelled data. 

With the objective of developing a versatile approach that performs decently on texts of different epochs without intense adaptation, Labusch et al.~\cite{labusch_bert_2019} experiment with BERT under different pre-training and fine-tuning settings. In a nutshell, they apply a model based on multilingual BERT embeddings, which is further pre-trained on large OCRed historical German unlabelled data (the Digital Collection of the Berlin State Library) and subsequently fine-tuned on several NE-labelled datasets (CoNLL-03, GermEval, and the German part of Europeana NER corpora). Tested across different contemporary/historical dataset pairs (similar to the all-corpora crossing of Riedl et al.~\cite{riedl_named_2018}), it appears that additional in-domain pre-training is most of the time beneficial for historical pairs, while performances worsen on contemporary ones. The combination of several task-specific training datasets has positive yet less important impact than BERT pre-training, as already observed by Schweter et al.~\cite{schweter_robust_2019}. Overall, this work shows that an appropriately pre-trained BERT model delivers decent recognition performances in a variety of settings. In order to further improve them, authors purpose to use the BERT large instead of the BERT base model, to build more historical labelled training data, and to improve the OCR quality of the collections.

The same spirit of combinatorial optimization drove the work of Todorov et al. \cite{todorov_transfer_2020} and Schweter et al.~\cite{schweter_triple_2020} in the context of HIPE-2020. Todorov et al. build on the bidirectional LSTM-CRF architecture of Lample et al. and introduce a multi-task approach by splitting the top layers for each entity type. Their general embedding layer combines a multitude of embeddings, on the level of characters, sub-words and words; some newly trained by the authors, as well as pre-trained BERT and HIPE's in-domain fastText embeddings. They also vary the segmentation of the input: line segmentation, document segmentation as well as sub-document segmentation for long documents. No additional NER training material was used for German and French, while for English, the Groningen Meaning Bank\footnote{\url{https://gmb.let.rug.nl/}} 
was adapted for training. Results suggest that splitting the top layers for each entity type is not beneficial. However, the addition of various embeddings improves the performance, as shown in the very detailed ablation test report. In this regard, character-level and BERT embeddings are particularly important, while in-domain embeddings contribute mainly to recall. Fine-tuning pre-trained embeddings did not prove beneficial. Using (sub-)document segmentation clearly improved results when compared to the line segmentation found in newspapers, emphasising once again the importance of context. Post-campaign F-scores for coarse literal NER are $75\%$ and $66\%$ for French and German (strict setting). English experiments yielded poor results, certainly due to the time and linguistic gaps between training and test data, and the pretty bad OCR quality of the material (in the same way as for \citet{provatorova_named_2020} and Kristanti et al.~\cite{kristanti_delft_2020}).

For their part, Schweter et al.~\cite{schweter_triple_2020} focused on German and experimented with ensembling different word and subword embeddings (modern fastText and historical self-trained and HIPE flair embeddings), as well as transformer-based language models (trained on modern and historical data), all integrated by the neural Flair NER tagging framework~\cite{akbik_flair_2019}. They used a state-of-the-art BiLSTM with an on-top CRF layer as proposed by \cite{huang_bidirectional_2015}, and perform sentence splitting and hyphen normalisation as pre-processing. To identify the optimal combination of embeddings and LMs, authors first selected the best embeddings for each type before combining them. Using richer representations (fastText<flair<BERT) leads to better results each time. Among the options, Wikipedia fastText embeddings proved better than the Common Crawl ones, suggesting that similar data (news) is more beneficial than larger data for static representations; HIPE flair embeddings proved better than other historical ones, likely because of their larger training data size and data proximity; and BERT LM trained on large data proved better than the one trained on historical (smaller) data. The best final combination includes fastText and BERT, leading to $65\%$ F-score on coarse literal NER (strict setting).

Finally, \citet{boros_robust_2020} also tackled NER tagging for HIPE-2020 in all languages and achieved best results. They used a hierarchical transformer-based model \cite{vaswani_attention_2017} built upon BERT in a multi-task learning setting. On top of the pre-trained BERT blocks (multilingual BERT for all languages, additionally Europeana BERT for German\footnote{\url{https://github.com/stefan-it/europeana-bert}} and  CamemBERT for French \cite{martin_camembert_2020}), two task-specific transformer layers were optionally added to alleviate data sparsity issues, for instance out-of-vocabulary words, spelling variations, or OCR errors in the HIPE dataset. A state-of-the-art CRF layer was added on top in order to model the context dependencies between entity tags. For base BERT with a limited context of 512 sub-tokens, documents are too long and newspaper lines are too short for proper contextualization. Therefore, an important pre-processing step  consisted in the reconstruction of hyphenated words and in sentence segmentation.  For the two languages with in-domain training data (French and German), their best run consisted in BERT fine-tuning, completed with the two stacked transformer blocks and the CRF layer. For English without in-domain training data, two options for fine-tuning were tested: a) training on monolingual CoNLL-03 data, and b) transfer learning by training on the French and German HIPE data. Both options worked better without transformer layers, and training on the French and German HIPE data led to better results. Final F-scores for coarse literal NER were $84\%$, $79\%$ and $63\%$ for French, German and English respectively (strict setting).

\vspace{0.15cm}
\noindent \textbf{Conclusion on deep learning approaches.}
What conclusions can be drawn from all this? First, the twenty or so papers reviewed above illustrate the growing interest of researchers and practitioners from different fields in the application of deep learning approaches to NER on historical collections. Second, it is obvious that these many publications also equate with a great diversity in terms of document, system and task settings. Apart from the historical German \cite{riedl_named_2018,schweter_robust_2019,labusch_bert_2019} and HIPE papers, most publications use different datasets and evaluation settings, which prevents result comparison; what is more, the sensitivity of DL approaches to experimental settings (pre-processing, embeddings, hyper-parameters, hardware) usually undermines any attempt to compare or reproduce experiments, and often leads to misconceptions about what works and what does not~\cite{yang_design_2018}. As shown in the DL literature review above, what is reported can sometimes be contradictory. However, and with this in mind, a few conclusions can be drawn:

\begin{itemize}[label={--}]
\setlength\itemsep{0.2em}
    \item State-of-the-art BiLSTM architectures achieve very good performances and largely outperform traditional CRFs, except in small data contexts and on very regular entities. As inference layer, CRF is a better choice than softmax (also confirmed by \citet{yang_design_2018}). Yet, in the fast-changing DL landscape, transformer-based networks are already taking over BiLSTM. 
    \item Character and sub-word information is beneficial and helps the model to deal with OOV words, presumably historical spelling variations and OCR errors. CNN appears to be a better option than LSTM to learn character embeddings.
    \item As for word representation, the richer the better. The same neural architecture performs better with character or word-based contextualised embeddings than with static ones, and even better with stacked embeddings. The combination of flair or fastText embeddings plus a BERT language model seems to provide an appropriate mix of morphological and lexical information. Contextualised representations also have positive impact in low resource setting.
    \item Pre-trained modern embeddings prove to transfer reasonably well to historical texts, even more when learned on very large textual data. As expected, in-domain embeddings contribute positively to performances most of the time, and the temporal proximity between the corpora from which embeddings are derived and the targeted historical material seems to play an important role. Although a combination of generic and historical prior knowledge is likely to increase performances, what is best between very large modern vs. in-domain LMs remains an open question.
    \item Careful pre-processing of input text (word de-hyphenation, sentence segmentation) in order to work with valid linguistic units appears to be a key factor.
\end{itemize}

\noindent Ultimately, apart from clearly outperforming traditional ML and rule-based systems, the most compelling aspect of DL approaches is certainly their transferability; if much still need to be investigated, the possibility of having systems performing (relatively) well across historical settings---or a subset thereof---seems to be an achievable goal.

\section{Strategies to deal with specific challenges}
\label{sec:strategies}

We report here on the main strategies implemented by the different types of approaches to overcome OCR noise, adapt to language shifts and deal with lack of resources. Table \ref{tab:challenges} provides a synthetic view of the challenges, their impact, and the possible solutions to address them.

\subsection{Dealing with noisy input}

There exist two main strategies to deal with OCR and OLR noise: adapting the input, i.e. correcting the text before parsing it, or adapting the tool, i.e. making the NER system capable of dealing with the noise. Let us recall here that OCR and OLR noise mostly correspond to: misrecognised characters, erroneously truncated or connected word parts, spurious word hyphenations and incorrect mix of textual segments, all of these translating into OOV words and/or inconsistencies affecting both entities and their context.

The first strategy corresponds to OCR/OLR post-correction and aims at recovering correct word forms and rebuilding linguistically motivated token sequences. Such processes depend on the specifics of each input (e.g. in terms of layout, typographic conventions, presence of marginalia) and are not easy to implement given the countless erroneous punctuation marks added by OCR and the subtle difference between soft and hard hyphens, an information often lost through the different digital versions of a document. In this context, most work apply a mix of well-known and ad hoc correction strategies based on corpus examination, including: a) correction of the `long s', an archaic from of the `s' letter systematically confused with `f' by OCR engines~\cite{alex_digitised_2012}; b) word de-hyphenation, which consists in removing the end-of-line soft hyphens and checking the validity of the resulting word form~\cite{alex_digitised_2012,dinarelli_treestructured_2012}; c) word OCR post-correction based on the edit distance between input tokens and a list of most frequent OOV words manually corrected (allowing to correct the majority of mistakes)~\cite{dinarelli_treestructured_2012}; d) application of a generic spelling correction system~\cite{huynh_when_2020}; and e) application of sentence segmentation~\cite{ortizsuarez_sinner_2020,boros_robust_2020}. Word de-hyphenation and sentence segmentation (or an approximation of it, depending on the quality of the input) are beneficial for all types of systems but are particularly critical for neural-based systems. Word correction have positive albeit moderate and irregular impact on the performances~\cite{dinarelli_treestructured_2012}, which illustrates the difficulty of OCR post-correction and raises the question of under which conditions it is most beneficial. Using a generic spelling correction system, Huynh et al.~\cite{huynh_when_2020} precisely leave aside the fine-tuning of OCR correction to focus, instead, on the question of when to apply it. They show that post-OCR correction is most beneficial with character and word-level error rates above 3\% and 20\% respectively, while it degrades NER performances for lower rates, due to spurious corrections. Overall, input noise correction or reduction is beneficial but should be adjusted according to the type and importance of noise.

\ra{1.15}
\begin{table*}[t] 
\footnotesize
\centering
\begin{tabular}{p{3.9cm}p{3.8cm}p{3.1cm}}
\multicolumn{1}{c}{\textbf{Challenges}}  & \multicolumn{1}{c}{\textbf{Impact}}  & \multicolumn{1}{c}{\textbf{Possible solutions}} \\
\midrule

\addlinespace[0.05cm]
\textbf{Noisy input} &  \multirow{4}{*}{\makecell[l]{sparser feature space,\\low recall.}}  & \multirow{3}{*}{\makecell[l]{OCR post-correction\\string similarity\\ historical LMs\\ in-domain training data\\ sub-word tokenisation}}\\
\tabitem OOV    &       & \\
\indent \tabitem broken token sequences   &   & \\
&&\\

\addlinespace[0.35cm]
\textbf{Dynamics of language}    &  \multirow{4}{*}{\makecell[l]{sparser feature space,\\low recall.} }  &  \multirow{4}{*}{\makecell[l]{normalisation\\ historical LMs\\ in-domain training data}}\\
\tabitem spelling variations       &      &  \\
\tabitem name irregularities            &   &  \\
\tabitem entity drift            &   &  \\

\addlinespace[0.35cm]
\textbf{Lack of resources} &  \multirow{4}{*}{\makecell[l]{limited learning capacities,\\limited system comparison.}}  &  \multirow{4}{*}{\makecell[l]{transfer learning\\active learning\\data augmentation\\resource sharing}}\\
\tabitem inappropriate typologies  &    &  \\
\tabitem lack of NE-annotated corpora   &   & \\
\tabitem paucity of historical LMs   &   & \\
\bottomrule
\end{tabular}
\caption{NER on historical documents: main challenges, their impact, and possible solutions.}
\label{tab:challenges}
\vspace{-6mm}
\end{table*}

Another approach is to leave the input untouched but to provide the system with relevant information about the noise, mainly in the form of embedded language representations for neural-based approaches.
In this regard, word embeddings computed at the character and sub-words levels (fastText~\cite{bojanowski_enriching_2017}) as well as character-level LM embeddings (flair~\cite{akbik_contextual_2018}) are particularly efficient in dealing with miss-recognised characters. For example, a BiLSTM-CRF model based on flair historical embeddings could correctly recognise and label \textit{T«i*louse} (Toulouse) and \textit{Caa.Qrs} (Cahors) as \textit{Location}, and \textit{o\r{}an} (\textit{Jean}) as \textit{Person}~\cite{bircher_toulouse_2019}. Beside input representations, transformer-based architectures such as BERT integrate sub-word tokenizers based on e.g. Byte-Pair Encoding~\cite{sennrich_neural_2016} or WordPiece~\cite{wu_google_2016} algorithms. Such tokenisation allows to consider word pieces not present in the LM vocabulary, thereby learning representations of OOV words. This is however not the total answer since resulting sub-words depend on the anatomy of the misspelling: if character insertion mostly results in known sub-word units, substitution or deletion produce uncommon units which still cause difficulties~\cite{sun_advbert_2020}. Boros et al.~\cite{boros_alleviating_2020} (outperforming their previous work \cite{boros_robust_2020}) carried out an in-depth error analysis of the performances of different variants of a BERT-based system augmented (or not) with extra transformer layers, considering among others the percentage of OOV words and the severeness of OCR noise in entity mentions. Conclusions are that the representation capacity of the extra layers is beneficial for both recall and precision: while misspelled entities are better recognised in general, the system with extra layers does not over-predict entities, as is the case for the non-augmented system. Authors also highlight the possible over-fitting of the base model on OCR-related patterns in frequent entities, a question on which further research is necessary.

\begin{figure*}[t]
        \centering
        \includegraphics[width=0.7\linewidth]{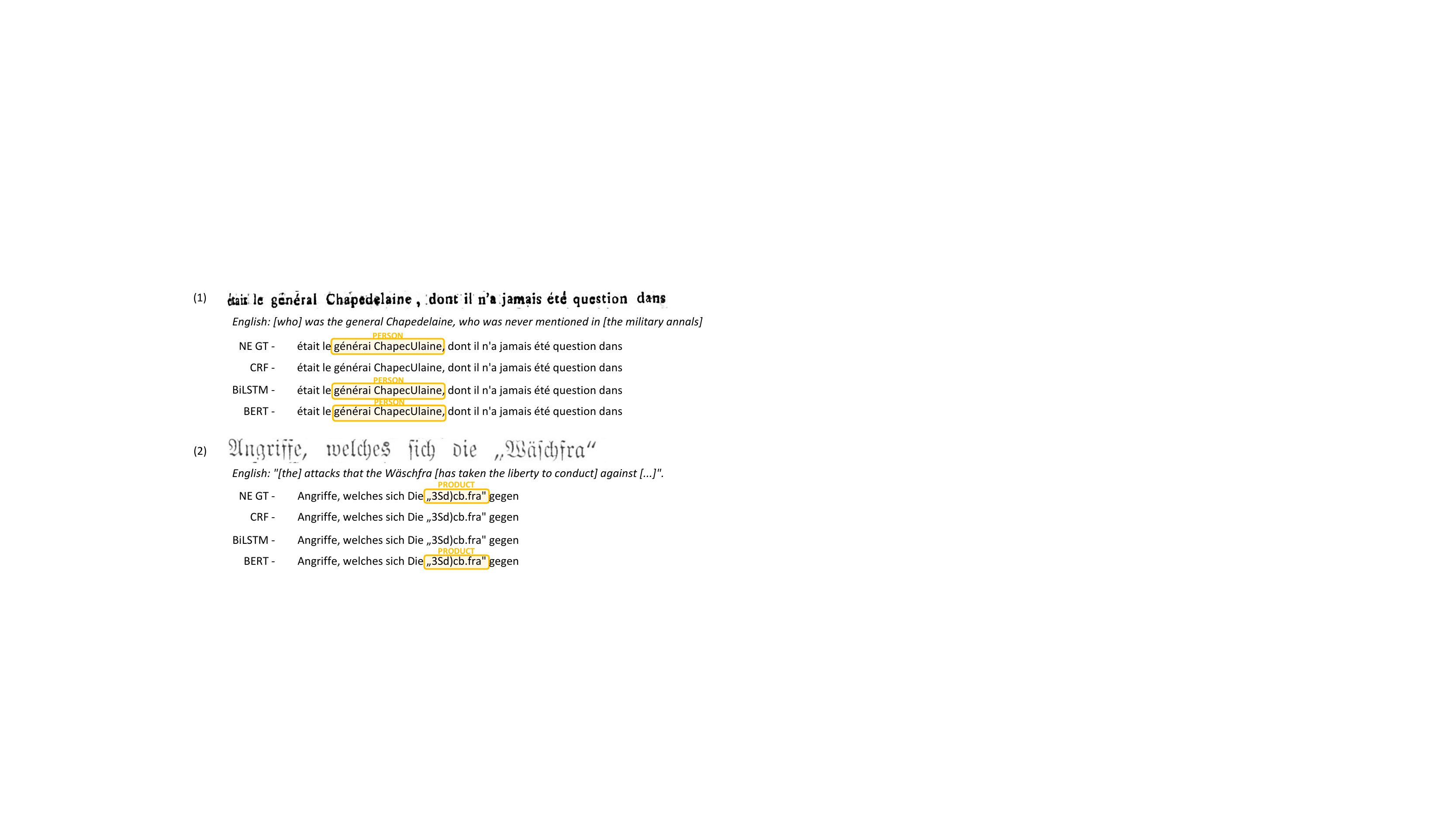}
        \caption{Results of a CRF, BiLSTM-CRF and BERT-based NER systems on excerpts from the French Swiss \textit{Gazette de Lausanne} of August 4 1818, p.4 (1), and from the German Luxembourgian \textit{Luxemburger Wort} of July 21 1868, p.2 (2) (HIPE data), compared to the ground truth (NE GT).}
        \label{fig:system-noise}
        \vspace{-2mm}
\end{figure*}

In general, language models and transformers thus appear as good options to deal with noisy input with the least effort, but the effectiveness of targeted correction heuristics should not be underrated. Let us conclude with Fig. \ref{fig:system-noise}, which illustrates the results of three different systems on noisy texts.

\subsection{Dealing with dynamics of language}
\label{sec:strategy-dia}

Similarly to noise, adaptation to dynamics of language aims at reducing the gap between the text to analyse and the knowledge of the system. Let us recall the issues at stake here: historical spelling variations, evolving naming conventions, and entity drift. There is no single and clear-cut answer to each of these issues, but rather a set of possibilities which can help address them.

As previously mentioned, rule-based and traditional ML systems using gazetteers often suffer from low recall due to vocabulary mismatches between name variants and gazetteer entries. Several approaches can help alleviate this issue. First, the lookup conditions can be loosened using a string similarity measure 
with an empirically determined threshold. In this respect, Borin et al.~\cite{borin_naming_2007} showed that allowing a small Levenshtein distance between input tokens and gazetteers entries allows to capture orthographic variations between 19C and 20C Swedish names and thus to increase the performances of their rule-based system, particularly recall (+5 percentage points). Another option consists in normalising historical spellings using transformation rules, either manually crafted, such as in Neudecker~\cite{neudecker_open_2016} and Platas et al.~\cite{platas_medieval_2020}, or automatically acquired on aligned texts such as in Kogkitsidou et al.~\cite{kogkitsidou_normalisation_2020}. Results obtained on normalised versions of texts or entities are usually better, though a somehow contrasted picture emerges depending on the system, the type of texts, and the time period --- as per OCR post-correction. 

When trying to adapt the knowledge of the system rather than the input, the key factor is, not surprisingly, temporal proximity.  Here word embeddings and language models derived from temporally close corpora seem to be better able to capture historical language idiosyncrasies, including spelling variations~\cite{schweter_robust_2019,labusch_bert_2019}. There is no clear evidence yet regarding what is best between pre-training from scratch on historical texts or fine-tuning a modern model on historical texts (in a supervised or unsupervised manner) and further research is necessary on this point. Beyond technicalities, an important aspect to consider when adapting a system to time is its application scope, i.e. whether it is intended to perform well on a unique target (one document type, one time period) or across several. 

\subsection{Dealing with the lack of resources}

As emphasised in Section \ref{sec:lack-resources}, the situation of lack of resources is not unique to historical NER and corresponds here to inappropriate typologies, and lack of labelled and unlabelled historical corpora. 

With respect to typologies, one can only adapt and/or define a typology when existing tag sets are not appropriate ~\cite{thompson_text_2016,ahmed_biofid_2019,platas_medieval_2020}. As easy as it may seem, two comments are in order here: first, this represents a time-consuming process which requires several expertise (in linguistics, in the historical domain at hands, and in knowledge representation) and needs to be documented, notably via annotation guidelines. Specifically, phenomena such as nested entities and metonymy did not received much attention in modern NER but are of high interest for humanists' (re)search needs. Second, careful attention should be paid to typologies interoperability, without which resources are mere silos and need an extra mapping step~\cite{rizzo_nerd_2012}. 

Several strategies can be adopted to cope with the lack of training data. The most widely used so far is transfer learning, as described in Section \ref{sec:DL}. Another option is active learning, where a ML system asks an oracle (or a user) to select the most relevant examples to consider, thereby lowering the number of data points required to learn a model. This is the approach adopted by Erdmann et al.~\cite{erdmann_challenges_2016} to recognise entities in various Latin classical texts, based on an active learning pipeline able to predict how many and which sentences need to be annotated to achieve a certain degree of accuracy, and later on released as toolkit to build custom NER models for the humanities~\cite{erdmann_practical_2019}.
Finally, another strategy is data augmentation, where an existing data set is expanded via the transformation of training instances without changing their label. This approach, which has not yet been deployed in a historical context, has been successfully tested by Dai et al.~\cite{dai_analysis_2020} on biomedical data, where several data augmentation techniques, in isolation or in combination, led to improved performance, especially with small training datasets.

\section{Conclusions and outlook}
\label{sec:ccl}


We presented an extensive survey of research, published in the last 20 years, on the topic of NER on historical documents. We introduced the main challenges of historical NER, namely document type and domain variety, noisy input, dynamics of language, and lack of resources. We inventoried existing resources available for historical NER (typologies and annotation guidelines, annotated corpora and language representations), and surveyed the approaches developed to date to tackle this task, paying special attention to how they adapt to historical settings. 

What clearly emerges from this review is, first, that research on historical NER has gained real momentum over the last decade. The availability of machine-readable historical texts coupled with the recent advances in deep learning has led to increased attention from researchers and cultural heritage practitioners for what has now become a cornerstone of semantic indexing of historical documents. Second, the body of research on historical NER started by following state-of-the-art approaches in modern NER (with rule-based and then traditional ML approaches), before fully experimenting with the various possibilities arising from diachronic transfer learning with neural-based approaches. This last development helped increase the performances of NER systems on historical material with F-scores going from 60-70\% on average for rule-based and traditional ML systems to, for the best neural systems, 80\%. As of today, it is therefore possible to design systems capable of dealing with historical and noisy inputs, whose performances almost compete with those obtained on contemporary texts. This success, however, should not conceal the progress still to be made. In this regard, we  outline a set of key priorities for the next generation of historical NER systems:
\begin{enumerate}
\setlength\itemsep{0.6em}
    \item \textit{Transferability}. We emphasise that beyond addressing a specific type of document and/or time period lies the question of systems' portability \textit{across} historical settings. While addressing system adaptability across both time and domain at once might be overly ambitious for the time being, having systems performing equally well across one or the other is highly desirable---especially for cultural heritage institutions---and represents to next great challenge. In this respect, we encourage to pursue and especially systematise further the transfer learning experiments undertaken so far.
    
    \item \textit{Robustness}. Although a great deal of headway has been made to enable systems to deal with atypical historical inputs, we highlight that OCR/OLR noise and historical spellings are still the main sources of errors of NER systems on historical texts. One of the way forward includes a better assessment of which type of noise is detrimental and to which extent in order to devise more systematic and focused strategies to deal with it. 
    
    \item \textit{System comparability}. The systematic comparison of the advantages and shortcomings of approaches to historical NER was made difficult because of the variety of settings to which they applied (domains, time periods, languages) and of the corpora against which they were evaluated. We stress the importance of gold standards and of shared tasks on historical material to enable system comparison and drive progress in historical NER.
    
    \item \textit{Finer-grained historical NER}. The (re)search interests of scholars go beyond the recognition of main entity types and we underline the need to carry out finer-grained NER, taking into account e.g. nested entities, entity name composition, and entity metonymy.
    
    \item \textit{Resource sharing}. All recent advances in deep learning were made possible by the availability of large-scale textual data. While the sharing of such resources has just begun, we emphasise the need for access to large-scale historical textual data or to language models derived thereof. This sharing should also extend to typologies, annotation guidelines, and training material, with a special attention to interoperability.
\end{enumerate}
\vspace{0.1cm}
\noindent NER on historical documents is an exciting field of research with high added-value for both NLP researchers and digital scholars. While the first can test the robustness of their approaches and gain new insights with respect to domain, language and time adaptation, the second can benefit from more accurate semantic indexing and text understanding of historical material. Lastly, we wish to mention two facets which, even if not directly related to development of historical NER systems, should be considered while working on this topic: any large-scale endeavour around historical NEs should acknowledge ethical and legal obligations related to personal data protection, and is most likely to be useful if humanities scholarship knowledge and needs are taken into account within an interdisciplinary framework.
\begin{acks}
The work of Maud Ehrmann and Matteo Romanello was supported by the Swiss National Science Foundation under the grants number CR-SII5\_173719 (\hyperlink{https://impresso-project.ch}{\textit{Impresso} - Media Monitoring of the Past}) and number PZ00P1\_186033 (only for MR). The work of Ahmed Hamdi, Elvys Linhares Pontes (now employed at the Trading Central Labs company) and Antoine Doucet was supported by the European Union's Horizon 2020 research and innovation program under grant 770299 (\hyperlink{https://www.newseye.eu/}{NewsEye}). 
\end{acks}

\bibliographystyle{ACM-Reference-Format}
\bibliography{bib/main}


\end{document}